\documentclass{article}

\usepackage{arxiv}

\usepackage[utf8]{inputenc} 
\usepackage[T1]{fontenc}    
\usepackage{hyperref}       
\usepackage{url}            
\usepackage{booktabs}       
\usepackage{amsfonts}       
\usepackage{nicefrac}       
\usepackage{microtype}      
\usepackage{lipsum}		
\usepackage{graphicx}
\usepackage{doi}
\usepackage{amsmath}
\usepackage{subfigure}
\usepackage{float}
\usepackage{subcaption}
\usepackage{graphicx}
\usepackage[numbers,square]{natbib}

\title{A STUDY OF UNIVERSAL ODE APPROACHES TO PREDICTING SOIL ORGANIC CARBON}

\author{ \href{https://orcid.org/https://orcid.org/0009-0007-0500-7147}{\includegraphics[scale=0.06]{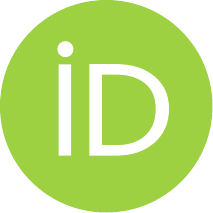}\hspace{1mm}Satyanarayana Raju G.V.V}\thanks{Use footnote for providing further
		information about author (webpage, alternative
		address)---\emph{not} for acknowledging funding agencies.} \\
	Human Sciences Research Center\\
	International Institute of Information Technology\\
	\texttt{gottumukkala.sa@research.iiit.ac.in} \\
        \And
	\href{https://orcid.org/0000-0000-0000-0000}{\includegraphics[scale=0.06]{orcid.pdf}\hspace{1mm} Prathamesh Dinesh Joshi} \\
	Vizuara AI Labs\\	
	\texttt{prathamesh@vizuara.com} \\
	\And
	\href{https://orcid.org/0000-0000-0000-0000}{\includegraphics[scale=0.06]{orcid.pdf}\hspace{1mm} Raj Abhijit Dandekar} \\
	Vizuara AI Labs\\
	Massachusetts Institute of Technology (prior) \\
	\texttt{raj@vizuara.com} \\
        \And
	\href{https://orcid.org/0000-0002-7025-6410}{\includegraphics[scale=0.06]{orcid.pdf}\hspace{1mm} Rajat Dandekar} \\
	Vizuara AI Labs\\
	Purdue University (prior) \\
    \texttt{rajatdandekar@vizuara.com} \\
        \And
	\href{https://orcid.org/0000-0001-6929-202X}{\includegraphics[scale=0.06]{orcid.pdf}\hspace{1mm}             Sreedhat      Panat} \\
	Vizuara AI Labs\\
	Massachusetts Institute of Technology (prior) \\
	\texttt{sreedhat@vizuara.com} \\
}

\renewcommand{\shorttitle}{\textit{arXiv} Template}

\hypersetup{
pdftitle={A STUDY OF UNIVERSAL ODE APPROACHES TO PREDICTING SOIL ORGANIC CARBON},
pdfsubject={q-bio.NC, q-bio.QM},
pdfauthor={Satyanarayana Raju G.V.V, Raj Dandekar, Rajat Dandekar, Sreedhat Panat},
pdfkeywords={First keyword, Second keyword, More},
}

\usepackage{graphicx}
\graphicspath{{figures/}}

\begin{document}
\maketitle
\begin{abstract}
Soil Organic Carbon (SOC) is a foundation of soil health and global climate resilience, yet its prediction remains difficult because of intricate physical, chemical, and biological processes. In this study, we explore a Scientific Machine Learning (SciML) framework built on Universal Differential Equations (UDEs) to forecast SOC dynamics across soil depth and time. UDEs blend mechanistic physics, such as advection--diffusion transport, with neural networks that learn nonlinear microbial production and respiration. Using synthetic datasets, we systematically evaluated six experimental cases, progressing from clean, noise-free benchmarks to stress tests with high (35\%) multiplicative, spatially correlated noise. Our results highlight both the potential and limitations of the approach. In noise-free and moderate-noise settings, the UDE accurately reconstructed SOC dynamics. In clean terminal profile at 50 years (Case 4) achieved near-perfect fidelity, with MSE = $1.6 \times 10^{-5}$, and $R^{2} = 0.9999$. Case~5, with 7\% noise, remained robust (MSE = $3.4 \times 10^{-6}$, $R^{2} = 0.99998$), capturing depth wise SOC trends while tolerating realistic measurement uncertainty. In contrast, Case~3 (35\% noise at $t=0$) showed clear evidence of overfitting: the model reproduced noisy inputs with high accuracy but lost generalization against the clean truth ($R^{2}=0.94$). Case~6 (35\% noise at $t=50$) collapsed toward overly smooth mean profiles, failing to capture depth wise variability and yielding negative $R^{2}$, underscoring the limits of standard training under severe uncertainty. Qualitatively, the UDE framework consistently preserved broad SOC patterns, avoided overfitting in moderate noise, and maintained physics-based plausibility even when data were corrupted. These findings suggest that UDEs are well-suited for scalable, noise-tolerant SOC forecasting, though advancing toward field deployment will require noise-aware loss functions, probabilistic modelling, and tighter integration of microbial dynamics.
\end{abstract}

\newpage
\section{Introduction}
Restoring soil health through improved management of Soil Organic Carbon (SOC) is a pressing challenge, particularly in tropical soils where degradation pressures are most severe \cite{bhattacharyya2011carbon}. Agricultural expansion and intensive cultivation have aggravated SOC loss, releasing large amounts of carbon into the atmosphere and intensifying the greenhouse effect. At the same time, SOC is fundamental for soil fertility\cite{gurmu2019soil}, influencing soil structure\cite{Jha2023}, nutrient cycling, and biological activity\cite{Xu2021} \cite{Ballantyne2018}, and is thus central to sustaining agricultural productivity\cite{Monti2024}. SOC levels are shaped by various factors such as climate, hydrology, parent material, fertility, vegetation, and land use \cite{ortner2022content}\cite{john2020using}, but they are also highly sensitive to human activities such as deforestation, biomass burning, and land use change. Globally, land conversion alone is estimated to transfer 1--2 petagrams of carbon each year from terrestrial ecosystems to the atmosphere, with 15--17\% of this flux attributed to SOC decomposition\cite{Singh2010}. 

These dynamics highlight the dual role of SOC in both environmental degradation and climate mitigation. The balance between carbon inputs and outputs in soil is a key factor that determines both the sequestration and long-term stability of soil organic carbon (SOC). Effective soil management, therefore requires not only an understanding of how SOC stocks are distributed across depths and landscapes, but also the ability to track and predict the dynamic carbon fluxes that shape them. These fluxes include processes such as carbon fixation, decomposition, release, and transformation within the soil system\cite{ding2025advancing}. Traditionally, SOC has been quantified through direct measurements such as soil core sampling. But these methods are often labor-intensive, time-consuming, and costly, limiting their scalability. This challenge highlights the growing importance of complementing field-based measurements with modeling approaches. Models provide a powerful tool to estimate SOC stocks and fluxes under varying land-use and management conditions, while also enabling projections of future changes driven by climate change. 

By integrating both direct measurements and simulation frameworks, researchers can more effectively assess SOC dynamics and design strategies that promote long-term carbon sequestration\cite{ding2025advancing}. In recent years, SOC management has gained renewed attention not only for improving soil productivity but also as a cornerstone of climate policy\cite{Li2024}. The dominant reasons for the persistence of SOC are stabilization processes, followed by repeated microbial processing of SOC\cite{Ahrens2015}. Enhancing SOC stocks contributes directly to carbon sequestration and is increasingly recognized within carbon markets as a tradable ecosystem service\cite{wang2024ensemble}. Reliable measurement and forecasting of SOC are therefore critical to support transparent carbon credit systems, offering both environmental and economic incentives for sustainable land management\cite{neofytou2024review}. 

\subsection{Soil Properties influencing Organic Carbon and Modelling}
Soil Organic Carbon (SOC) is a critical component for maintaining soil health and agricultural productivity, as it influences various soil properties and functions\cite{Kakhani2024} \cite{john2020using}. A combination of natural factors and human-induced land management practices shapes its levels \cite{Berardi2024}. SOC is fundamental for soil fertility and quality, acting as a pivotal element for terrestrial ecosystems and having significant implications for food security and agricultural sustainability\cite{luo2017soil} \cite{bursac2022instance}. It is considered a key indicator of soil health, which the Intergovernmental Technical Panel on Soils (ITPS) defines as the soil's ability to sustain the productivity, diversity, and environmental services of terrestrial ecosystems\cite{ermolieva2024tracking}. Specifically, SOC contributes to soil health and productivity through soil structure, leading to increased structural stability. SOC is crucial for nutrient absorption by plants and overall soil fertility. It also improves the chemical properties of soil\cite{neofytou2024review}. 

While not explicitly stated as 'biological activity', SOC is essential for allowing soils to provide services that sustain the diversity of terrestrial ecosystems, making it a key indicator of soil health. SOC improves the biological properties of soil\cite{ermolieva2024tracking}. SOC positively impacts soil water retention, water-holding capacity, and permeability \cite{neofytou2024review}. High SOC levels correlate with high plant productivity and increased crop yields, reducing the need for chemical fertilizers. It directly improves soil productivity\cite{pavlovic2024deep}. 

Soil Organic Carbon (SOC) content is influenced by a variety of soil properties, including pH, cation exchange capacity (CEC), and clay content\cite{kusmierz2023soil} \cite{makipaa2024organic} \cite{solly2020critical}. These factors control SOC dynamics through mechanisms such as stabilization on mineral surfaces, formation of organo-mineral associations, and influencing microbial activity\cite{Zech2024}. The interplay between these properties determines the amount and stability of SOC in different soil types and under various land uses \cite{rabot2024relevance}. Soil pH is a major property that discriminates between different soil types and influences SOC stabilization \cite{rabot2024relevance}. In a study of French soils, the SOC/clay ratio was found to be strongly affected by soil pH. Acidic soils (pH $<$ 5) consistently had high SOC/clay ratios and were classified as healthy, while alkaline soils (pH $>$ 8) often had low ratios and were classified as unhealthy \cite{rabot2024relevance}.  In Swiss forest soils, the relationship between SOC and effective CEC (CEC\textsubscript{eff}) is mediated by pH\cite{solly2020critical}. In acidic soils, particularly those with high mean annual precipitation, the presence of Al\textsuperscript{3+} cations is common. These cations are associated with the formation of organo-metal complexes, which can stabilize SOC \cite{solly2020critical}. Effective cation exchange capacity (CEC\textsubscript{eff}) can be causally linked to SOC preservation because it reflects the reactive soil surfaces available for SOC adsorption and represents a direct measure of SOC sorption at existing pH conditions \cite{solly2020critical}. In Burundi soils, CEC is influenced by soil pH, clay content, and organic carbon content, with organic carbon emerging as a key determinant of CEC in these highly weathered, kaolinitic soils \cite{kaboneka2024predicting}.  Clay content is a significant abiotic factor influencing SOC levels, particularly at a regional scale. This is attributed to clay's high specific surface area and its ability to form stable micro-aggregates, which protect SOC. 

While some studies show positive correlations between clay and SOC, others suggest that considering only clay content can oversimplify SOC preservation mechanisms, as mineralogy and other factors are also crucial \cite{solly2020critical}. In a Brazilian study, clay was the most prominent soil property influencing SOC stocks, with its relative importance increasing with soil depth \cite{luo2017soil}. pH, CEC, and clay content all exert significant control over soil organic carbon, but their effects are interconnected and context-dependent.

With increasing data availability, more complex Deep Learning (DL) approaches \cite{Lusch2018}\cite{timm2006neural} are being used for SOC prediction, as they can potentially capture more complex non-linear relationships than traditional ML algorithms. DL models, particularly SSL-SoilNet, have demonstrated high accuracy in SOC prediction\cite{Kakhani2024}. Specific DL methods mentioned include Convolutional Neural Networks, Recurrent Neural Networks (RNN), and Long Short-Term Memory (LSTM) models\cite{wang2023comparison}. DL algorithms have generally outperformed Random Forest for SOC prediction, with LSTM models often performing the best among DL approaches\cite{wang2023comparison}. 

Also, there are traditional models such as RothC \cite{Metrikaityte2025} \cite{paul2020evaluation}, which simulate SOC dynamics based on ecological processes. Both RothC and Century are considered robust foundational models because they establish strong, balanced bidirectional linkages between theory, measurement, and modeling in evaluating soil organic matter dynamics\cite{schimel2023modeling}\cite{paul2020evaluation}. However, there is a growing demand for new generations of more mechanistically realistic models that can better describe existing datasets related to fluxes, pool sizes, and organic matter chemistry. While traditional models are valuable, these approaches struggle to represent the nonlinear interactions between soil health factors (pH, organic matter, cation exchange capacity, clay content), climate, management practices, and biological interactions \cite{Berardi2024}. 

To address these limitations, \textbf{Scientific Machine Learning (SciML)} offers a hybrid framework that blends physical principles with data-driven learning \cite{rackauckas2020universal} \cite{raissi2019pinn} \cite{bolibar2023glacier} \cite{chen2018neuralODE} \cite{dandekar2022phd} \cite{baker2019workshop} \cite{Verma2024} \cite{Cuomo2022}. In particular, Universal Differential Equations (UDEs) embed mechanistic partial differential equations (PDEs) within neural networks capable of approximating complex or unobserved processes \cite{rackauckas2020universal} \cite{cai2021pinnfluid} \cite{karniadakis2021pinnreview} \cite{Hossain2024} \cite{Bonnaffe2021}. This approach allows known soil physics to be preserved while enabling the model to learn nonlinear biological dynamics from data. With the twenty-first century expected to bring some of the most rapid climatic shifts in Earth’s history, and given that soil carbon fluxes—closely linked to microbial activity—play a central role in regulating greenhouse gases, advancing our capacity to represent microbial processes has become crucial \cite{Singh2010}. Within the SOC UDE framework, this means developing models that explicitly capture microbial contributions to carbon stabilization and release, thereby improving predictions of soil carbon dynamics under accelerating climate change. 

In this study, we develop a UDE model for SOC forecasting in which neural networks approximate production and respiration processes as functions of climatic variables, land use practices, soil health indicators, and biological interactions. By integrating scientific knowledge with machine learning, we aim to improve the mathematical representation of microbial influences on SOC and predictive accuracy.

This paper is structured as follows. We begin with the mathematical formulation of Soil Organic Carbon (SOC) dynamics, where the known physics of advection–diffusion–reaction are coupled with neural networks to capture unknown production and respiration processes. We then describe the methodology and training strategy, including how synthetic soil datasets were generated, how noise was added to mimic realistic measurement uncertainty, and how the UDE models were trained and optimized. To test the robustness of the framework, we design six experimental cases: starting with a clean baseline, then progressively adding noise to drivers and terminal SOC values under moderate and high uncertainty scenarios. The results section presents the outcomes of these experiments, with a focus on hyperparameter tuning, prediction accuracy, residual behavior, and visual comparisons of the true and modeled SOC profiles. Building on these findings, we move into a discussion and conclusion, reflecting on the strengths and limitations of the UDE approach for SOC prediction. Finally, we point to future research directions, such as using symbolic regression for better interpretability, incorporating microbial and climatic drivers, and scaling the models toward probabilistic and spatially explicit SOC–climate interactions.

\section{Mathematical Formulation and Methodology}
The core objective of this study is to develop a physics-guided hybrid modeling framework that predicts the temporal and spatial evolution of Soil Organic Carbon (SOC) using a Universal Differential Equation (UDE) formulation. The model embeds neural networks within a Partial Differential Equation (PDE) to capture complex, nonlinear biological processes \cite{Bonnaffe2021} \cite{Monti2024} \cite{vogel2024adequately} such as microbial respiration and carbon input from organic sources, modulated by climatic factors. The PDE represents known physical transport processes (diffusion and advection), while the embedded neural networks learn unknown microbial dynamics using soil health indicators (pH, CEC, Clay content), and organic carbon (OC) profiles, which are provided as data inputs for supervised learning.

\subsection*{Neural ODEs and Universal Differential Equations}
Differential equations are fundamental tools for modeling dynamical systems in 
science and engineering. Recently, Scientific Machine Learning (SciML) has 
introduced ways to combine these mechanistic models with the flexibility of neural networks. Two key approaches in this domain are 
\textbf{Neural Ordinary Differential Equations (Neural ODEs)} and 
\textbf{Universal Differential Equations (UDEs)}.

Neural ODEs, introduced by \citet{Chen2018}, replace the traditional discrete layers of a neural network with a continuous transformation defined by an ODE. In practice, this means that instead of stacking layers, the dynamics of hidden states are modeled as
\[
\frac{dh(t)}{dt} = f(h(t), t; \theta),
\]
where $f$ is a neural network parameterized by $\theta$. Solving this ODE with a numerical integrator yields the output of the model. Neural ODEs are powerful because they provide memory-efficient training, adapt their depth dynamically, and naturally model time-continuous processes. They have been successfully applied in physics, biology, and control systems where system dynamics are important \citep{Kidger2022}.

Universal Differential Equations (UDEs), proposed by \citet{rackauckas2020universal}, 
extend this idea by embedding neural networks directly into mechanistic 
differential equations. Instead of replacing the entire right-hand side with a neural function, UDEs keep the known physics-based components (e.g., diffusion, advection) and only learn the unknown or poorly understood parts (e.g., complex biological interactions). Formally, this can be written as:
\[
\frac{du}{dt} = f_{\text{physics}}(u,t) + f_{\text{NN}}(u,t;\theta).
\]
This hybrid approach ensures that models respect established physical laws while 
leveraging data-driven learning for nonlinear or uncertain processes. For 
example, in soil carbon modeling, diffusion and transport terms can be defined mechanistically, while microbial production and respiration dynamics are learned from data. Together, Neural ODEs and UDEs exemplify the strength of SciML: they balance mechanistic interpretability with the flexibility of machine learning, offering a pathway to more accurate and generalizable models in data-limited scientific domains.

\subsection*{\textbf{Governing Equations}}
Using the predicted concentration profile for the constant coefficient advection-diffusion equation\cite{zoppou1997analytical}, we consider the SOC concentration $C(z,t)$ as a function of depth $z$ and time $t$. Here, SOC (or) OC (or) C are used interchangeably. The dynamics are described by (Using one-dimensional time-dependent Burgers’ equation and adding Neural Network (NN)
terms) \cite{jiwari2015hybrid}:
\begin{equation}
\frac{\partial C}{\partial t} = D \frac{\partial^2 C}{\partial z^2} - v \frac{\partial C}{\partial z} + P(C,\mathbf{env}) - R(C,\mathbf{env}),
\label{eq:governing}
\end{equation}

where:
\begin{itemize}
    \item $D$ = diffusion coefficient,
    \item $v$ = advection (vertical transport) velocity,
    \item $P(C,\mathbf{env})$ = production term (biological carbon inputs),
    \item $R(C,\mathbf{env})$ = respiration term (biological carbon losses),
    \item $\mathbf{env} = \{ \text{pH}, CEC, Claycontent\}$ = soil environment variables representing soil health parameters, env(z, t) as a function of depth and time.
\end{itemize}

\textbf{Neural Network Representation of Biological Terms}
The production and respiration processes are highly nonlinear and driven by soil properties and biological interactions. We approximate these using two neural networks:
\begin{align}
P(C,\mathbf{env}) &\approx \text{NN}_P(\text{pH}, OC, CEC, Claycontent, z, t; \theta_P), \\
R(C,\mathbf{env}) &\approx \text{NN}_R(\text{pH}, OC, CEC, Claycontent, z, t; \theta_R),
\end{align}
where $\theta_P, \theta_R$ are the trainable parameters of the respective networks.

Thus, the UDE becomes\cite{jiwari2015hybrid}:
\begin{equation}
\frac{\partial C}{\partial t} = D \frac{\partial^2 C}{\partial z^2} - v \frac{\partial C}{\partial z} + \text{NN}_P(C,\mathbf{env}) - \text{NN}_R(C,\mathbf{env}).
\label{eq:ude}
\end{equation}

\textbf{Boundary and Initial Conditions}
\begin{align}
C(z,0) &= C_0(z), \\
\frac{\partial C}{\partial z}\bigg|_{z=0,H} &= 0 \quad \text{(No-flux boundary conditions\cite{zoppou1997analytical})}.
\end{align}

\subsection{Numerical Implementation Steps}

\begin{enumerate}
    \item \textbf{Discretize soil depth}: Divide the soil profile into $N$ layers $\{z_1,z_2,\dots,z_N\}$.
    \item \textbf{Generate synthetic soil property profiles}: Create synthetic values for pH, OC, CEC and Clay content for each depth and time.
    \item \textbf{Define Neural Networks}:
    \begin{itemize}
        \item $\text{NN}_P$: Predicts production from $[\text{pH}, OC, CEC, Clay content, z, t]$.
        \item $\text{NN}_R$: Predicts respiration from $[\text{pH}, OC, CEC, Clay content, z, t]$.
    \end{itemize}
    \item \textbf{Combine PDE + NNs into UDE} as shown in Equation~\eqref{eq:ude}.
    \item \textbf{Generate synthetic target data}: Simulate using zero NN outputs (Production and Respiration are zero) for baseline comparison.
    \item \textbf{Loss Function}: Mean Squared Error (MSE) \cite{jadon2024regressionloss} between predicted and synthetic SOC profiles:
    \begin{equation}
    \mathcal{L} = \frac{1}{N_T N_Z} \sum_{i=1}^{N_T} \sum_{j=1}^{N_Z} \big( OC_{\text{pred}}(z_j,t_i) - OC_{\text{target}}(z_j,t_i) \big)^2.
    \end{equation}
    \item \textbf{Optimization}: Train $\theta_P,\theta_R$ using ADAM + BFGS.
\end{enumerate}

\subsection{Methodology and Training Strategy}
In this study, we model Soil Organic Carbon (SOC) dynamics using a discretized advection–diffusion-reaction partial differential equation (PDE). The advection and diffusion terms capture physics-based transport processes, such as vertical mixing and movement of carbon through the soil profile, while the reaction term represents complex biological processes such as carbon production and respiration. Unlike traditional models that use fixed empirical functions for these processes, we embed neural networks to learn the unknown source and sink terms directly from data. The PDE is numerically integrated forward in time for a 50-year horizon, producing predictions of SOC evolution under given soil and environmental inputs. 

Training is achieved by differentiating through the solver using adjoint sensitivity analysis, which allows the neural networks to be optimized efficiently with respect to both the observational data and the governing PDE constraints. To improve robustness, we employ specialized loss handling techniques that manage occasional solver failures (e.g., divergence or instability during training), ensuring stable learning across a wide range of hyperparameter settings. Model performance is further enhanced through systematic hyperparameter tuning. We perform grid searches over neural network architectures (depth, width, activation functions), optimizer settings, and loss weightings that balance the influence of data fitting against PDE consistency. This combination of mechanistic modeling and data-driven learning enables the framework to capture both well-understood physical processes and nonlinear biological dynamics that are otherwise difficult to specify explicitly.\newline

\subsection*{Synthetic Training Dataset}
To evaluate and validate the Universal Differential Equation (UDE) approach for soil organic carbon (SOC) modeling, a synthetic dataset has been generated. This dataset mimics realistic soil conditions by specifying smooth depth- and time-dependent profiles for key soil drivers. The following variables were included:
\begin{itemize}
    \item Soil Organic Carbon (SOC), denoted as $C(z,t)$
    \item Soil pH, denoted as $\mathrm{pH}(z,t)$
    \item Cation Exchange Capacity (CEC), denoted as $\mathrm{CEC}(z,t)$
    \item Clay content, denoted as $\mathrm{Clay}(z,t)$
\end{itemize}
The spatial domain is defined as depth $z \in [0,1]$~m, discretized into $N_z=30$ grid points. The temporal domain is $t \in [0,50]$ years.

\paragraph{Initial SOC profile:}
The synthetic initial condition at $t=0$ is specified as:
\begin{equation}
    C(z,0) = C_0 \, \exp(-k_{\text{decay}} \, z \, L),
\end{equation}
where $C_0 = 1.2$ is the surface SOC concentration, $k_{\text{decay}} = 0.02$, and $L=100$ is a depth scaling factor. This generates a decreasing exponential SOC profile with depth.

\paragraph{Soil pH:}
The pH varies smoothly with depth and time:
\begin{equation}
    \mathrm{pH}(z,t) = 6.5 - 0.5\,z + 0.10 \,\sin\!\left(\frac{2\pi t}{1.0}\right).
\end{equation}
This captures both depth-dependent acidity and seasonal oscillations.

\paragraph{Cation Exchange Capacity (CEC):}
CEC is modeled as a function of depth and long-term oscillations:
\begin{equation}
    \mathrm{CEC}(z,t) = 0.5 + 0.1 \,\sin(2\pi z) + 0.05 \,\cos\!\left(\frac{2\pi t}{5.0}\right).
\end{equation}

\paragraph{Clay content:}
Clay content varies spatially with depth and shows slow temporal fluctuations:
\begin{equation}
    \mathrm{Clay}(z,t) = 25.0 + 5.0 \,\cos(2\pi z) + 0.5 \,\sin\!\left(\frac{2\pi t}{10.0}\right).
\end{equation}

Together, these synthetic functions define a controlled and fully specified training dataset:
\begin{equation}
    \mathcal{D} = \{ (z,t) \mapsto (C(z,t), \mathrm{pH}(z,t), \mathrm{CEC}(z,t), \mathrm{Clay}(z,t)) \},
\end{equation}
which provides both the target SOC profile and its drivers for UDE training and benchmarking. The pH, CEC, Clay content denote the soil environment input $\text{env}(z, t)$ to the neural networks.

\subsubsection*{Noise Models for Soil Health Parameters:}
To evaluate model robustness under realistic uncertainties, we added two types of noise to the synthetic soil health parameters ($\mathrm{pH}$, $\mathrm{CEC}$, and clay content): multiplicative Gaussian noise \cite{sancho1982multiplicative} and spatially correlated noise. Multiplicative noise models the fact that measurement errors in soil properties often scale with the magnitude of the quantity being measured. For a soil parameter $x(z,t)$, the noisy observation $\tilde{x}(z,t)$ is \cite{sancho1982multiplicative} defined as:

\begin{equation}
\tilde{x}(z,t) \;=\; x(z,t) \cdot \big( 1 + \eta \cdot \epsilon(z,t) \big),
\end{equation}

where $\epsilon(z,t) \sim \mathcal{N}(0,1)$ is a standard Gaussian random variable and $\eta$ is the relative noise level (e.g., $\eta = 0.07$ for $7\%$ noise). This formulation ensures that larger values of $x(z,t)$ are perturbed more strongly, reflecting proportional uncertainty in field or lab measurements. 

\subsection*{Training Strategy}
The Production term representing the carbon inputs and the respiration term representing the carbon losses, affected by land use, climatic factors, elevation, slope parameters etc., two neural networks are defined as
\begin{align}
P(C,\mathbf{env}) &\approx \text{NN}_P(\text{pH}, OC, CEC, Claycontent, z, t; \theta_P), \\
R(C,\mathbf{env}) &\approx \text{NN}_R(\text{pH}, OC, CEC, Claycontent, z, t; \theta_R),
\end{align}
where $\theta_P, \theta_R$ are the trainable parameters of the respective networks. Each network uses a fully connected feedforward architecture with nonlinear activation functions (e.g., $\tanh$ or $GeLU$). We train end-to-end by differentiating through the ODE solver with adjoint automatic differentiation and enforce numerical stability with bounded outputs and safe loss penalties.

\subsection*{Discretization and Integration}
The PDE is discretized by the method of lines:
\begin{itemize}
  \item Depth is discretized with $N_z = 30$ uniformly spaced points on $[0,1]$, spacing $\Delta z$.
  \item Interior derivatives use second-order central differences:
  \[
  u_{zz}(z_i,t) \approx \frac{u_{i+1}-2u_i+u_{i-1}}{\Delta z^2}, \quad
  u_{z}(z_i,t) \approx \frac{u_{i+1}-u_{i-1}}{2\Delta z}.
  \]
  \item Zero-flux (Neumann) boundary conditions are imposed by duplicating interior slopes.
\end{itemize}

This yields a system of $N_z$ ODEs, which we integrate with the Tsitouras 5th-order explicit method (\texttt{Tsit5()}). During hyperparameter tuning, solver tolerances are set to $10^{-5}$; during final evaluation they are tightened to $10^{-6}$. The right-hand side (RHS) of the PDE is implemented in a non-mutating, out-of-place form to ensure safe differentiation with Zygote.

\subsection*{Training Objective and Differentiation}
We train against the \emph{terminal-only} loss for say t = 50 years using MSE\cite{jadon2024regressionloss}:
\begin{equation}
\mathcal{L}(\theta) = \frac{1}{N_z} \sum_{i=1}^{N_z} \big( u_\theta(z_i,50) - u_{\text{data}}(z_i,50) \big)^2,
\end{equation}
where $u_\theta$ is the model-predicted SOC profile at $t=50$.

Gradients are obtained by differentiating through the ODE solver with the adjoint method:
\texttt{InterpolatingAdjoint(autojacvec=ZygoteVJP())}.
This approach is memory-efficient and leverages Zygote’s vector-Jacobian products. Optimization uses the Adam optimizer.

\subsection*{Robustness to Solver Failures}
NaNs or solver divergence during training can cause undefined gradients.  
We ensure robustness through:
\begin{enumerate}
  \item A \texttt{safe\_solve} wrapper: catches solver errors and returns a controlled penalty.
  \item A fallback penalty loss:
  \[
  \mathcal{L}_{\text{fail}}(\theta) = 10^6 + 0 \cdot \|\theta\|^2,
  \]
  which discourages unstable regions but still depends on parameters so gradients exist.
\end{enumerate}
This avoids the common \texttt{iterate(::Nothing)} error and allows training to proceed smoothly.

\subsection*{Hyperparameter Tuning}
A compact grid search explores:

\begin{table}[h]
\centering
\caption{Hyperparameter grid for UDE training}
\begin{tabular}{ll}
\hline
\textbf{Hyperparameter} & \textbf{Values explored} \\
\hline
Hidden sizes ($H_1, H_2$) & $H_1 \in \{32,64\}, \ H_2 \in \{16,32\}$ \\
Activation functions      & $\tanh$, GeLU \\
Learning rates            & $3 \times 10^{-3}, \ 5 \times 10^{-3}$ \\
\hline
\end{tabular}
\end{table}

The best configuration is selected by terminal loss.

\subsection*{Final Training and Evaluation}
The best hyperparameters are retrained with tighter solver tolerances. Final evaluation integrates forward without adjoints (no gradient calculation needed), providing stable predictions.

Results are visualized via:
\begin{itemize}
  \item Predicted vs true SOC depth profiles,
  \item Residuals vs depth,
  \item A heatmap displaying \texttt{[True | Pred | Residual]}.
\end{itemize}

\subsection*{Implementation Choices for Stability and Speed}
\begin{itemize}
  \item All computations use \texttt{Float32} for speed.
  \item Out-of-place RHS and no static arrays improve autodiff stability.
  \item Non-negative outputs enforce biological plausibility.
  \item Early stopping accelerates hyperparameter tuning.
\end{itemize}

\subsection*{Computational Cost}
Each training iteration requires one forward solve and one adjoint solve to $t=50$. With $N_z=30$ and shallow networks, runs are tractable. Early stopping makes the grid search efficient.

\subsection*{Reproducibility}
\begin{itemize}
  \item Packages: \texttt{Lux}, \texttt{DifferentialEquations}, \texttt{SciMLSensitivity}, \texttt{Zygote}, \texttt{Optimization}, \texttt{Plots}.
  \item Random seeds are fixed per trial for reproducibility.
  \item Solver tolerances are reported explicitly.
\end{itemize}

\section{Experimental Cases for UDE-Based SOC Modeling}
We designed six experiments using a synthetic Soil Organic Carbon (SOC) dataset, with soil parameters (OC, pH, CEC, Clay) over depth--time $(z,t)$, to test a Universal Differential Equation (UDE) framework. Each case combines a time focus (initial profile at $t=0$ vs. terminal profile at $t=50$ years) with different levels of noise (none, moderate, high) applied to both inputs and outputs. All training runs start from random initialization, and hyperparameters (network depth/width, learning rate, loss weights, etc.) are optimized for each case.  

\subsection*{Case 1: Hyperparameter Tuning at $t=0$ (Initial SOC), No Noise (Baseline)}
\begin{figure}[htbp]
    \centering
    \begin{subfigure}
        \centering        \includegraphics[width=0.6\linewidth]{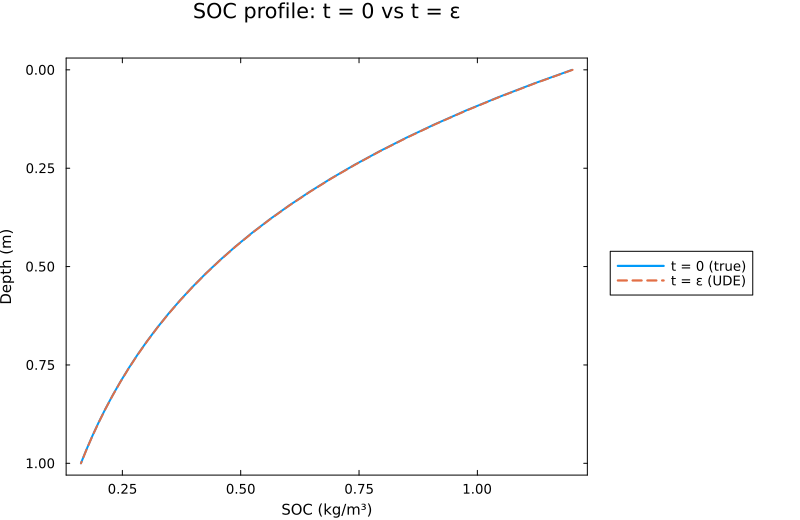}
        \caption{True vs. UDE prediction}
        \label{fig:True_vs_UDE-(SOC)-Case-1}
    \end{subfigure}
    \hfill
    \begin{subfigure}
        \centering        \includegraphics[width=0.6\linewidth]{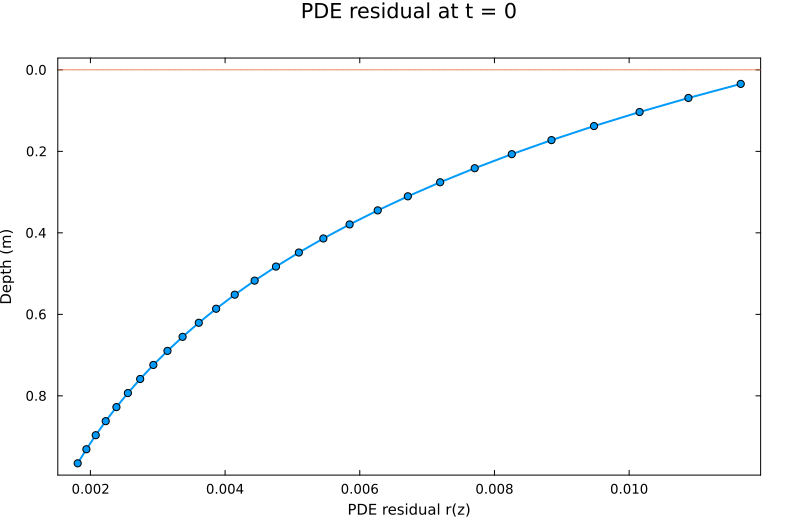}
        \caption{Residuals distribution}
        \label{fig:Residual-Distribution-Case-1}
    \end{subfigure}
    \vspace{0.8cm}
    \begin{subfigure}
        \centering        \includegraphics[width=0.6\linewidth]{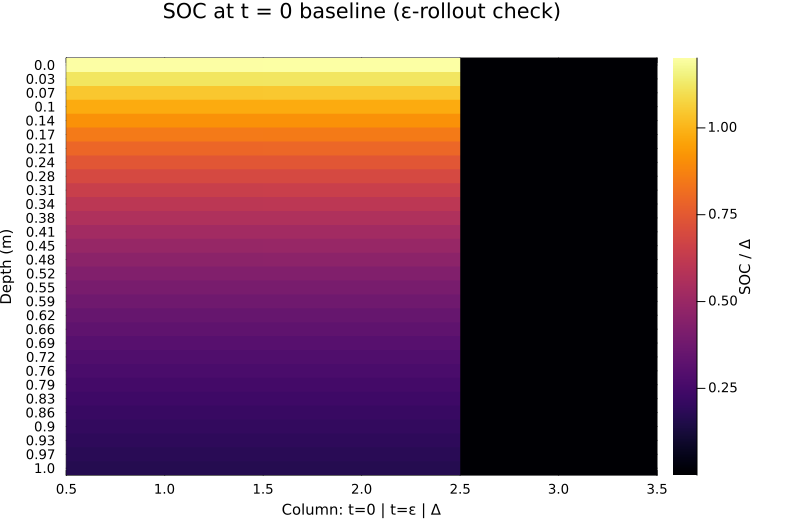}
        \caption{Heatmap of True, UDE, and Residual}
        \label{fig:Heatmap-Case-1}
    \end{subfigure}    
    \caption*{Case 1 results: (a) comparison of True vs. UDE predictions, 
             (b) residual error plot, and (c) heatmap of True SOC, UDE-predicted SOC, 
             and residual differences.}

\end{figure}
Universal Differential Equation (UDE) results for Soil Organic Carbon (SOC) dynamics under baseline, noise-free conditions. In this first case, we trained the UDE on a clean synthetic dataset generated from the advection–diffusion–reaction model of SOC, using soil health parameters (pH, CEC, clay content, and organic carbon inputs) at t=0 years. Because no noise was added, this setting represents the “best-case” scenario for learning SOC behavior. Panel (a) Figure~\ref{fig:True_vs_UDE-(SOC)-Case-1} shows the comparison between the true synthetic SOC profile and the UDE-predicted SOC profile across soil depth. The agreement is almost exact: the UDE smoothly follows the depthwise variations of SOC, demonstrating that the mechanistic terms (advection and diffusion) together with the neural network component can fully reconstruct the baseline distribution. Panel (b) Figure~\ref{fig:Residual-Distribution-Case-1} presents the residual map, highlighting the difference between predicted and ground truth values. The residuals are nearly negligible across all depths, confirming that the UDE fit respects both the data and the governing equations. Panel (c) Figure~\ref{fig:Heatmap-Case-1} displays the heatmaps of the true SOC, the UDE prediction, and their residuals. This visualization reinforces the accuracy of the model under noise-free conditions: the true and predicted SOC heatmaps overlap almost perfectly, while the residual heatmap remains close to zero across the domain. Case 1 demonstrates that under ideal conditions, the UDE framework can replicate the synthetic SOC profile with high fidelity. This establishes a performance ceiling against which the robustness of the model under noisy conditions (Cases 2,3,5,6) can later be assessed.

\subsection*{Case 2: Hyperparameter Tuning at $t=0$, Moderate Noise (7\%)}
\begin{figure}[htbp]
    \centering
    \begin{subfigure}
        \centering        \includegraphics[width=0.6\linewidth]{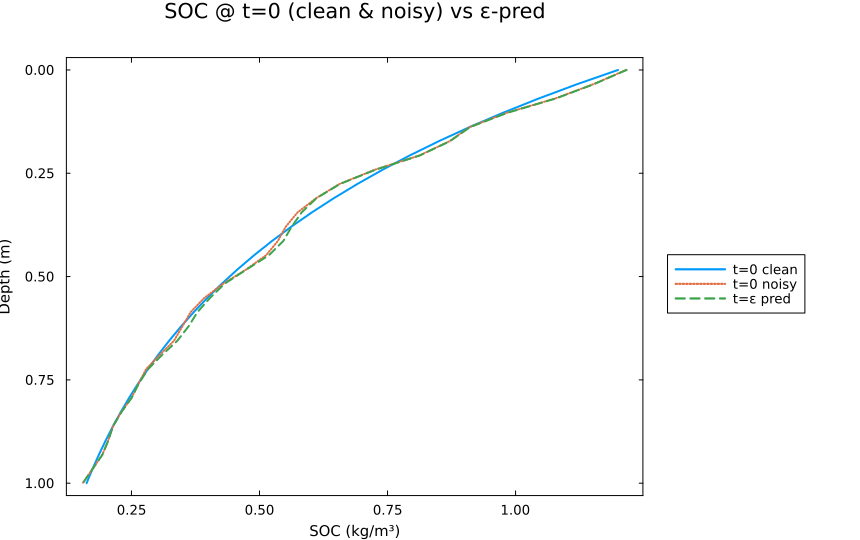}
        \caption{True vs. UDE prediction}
        \label{fig:True_vs_UDE-(SOC)-Case-2}
    \end{subfigure}
    \hfill
    \begin{subfigure}
        \centering        \includegraphics[width=0.6\linewidth]{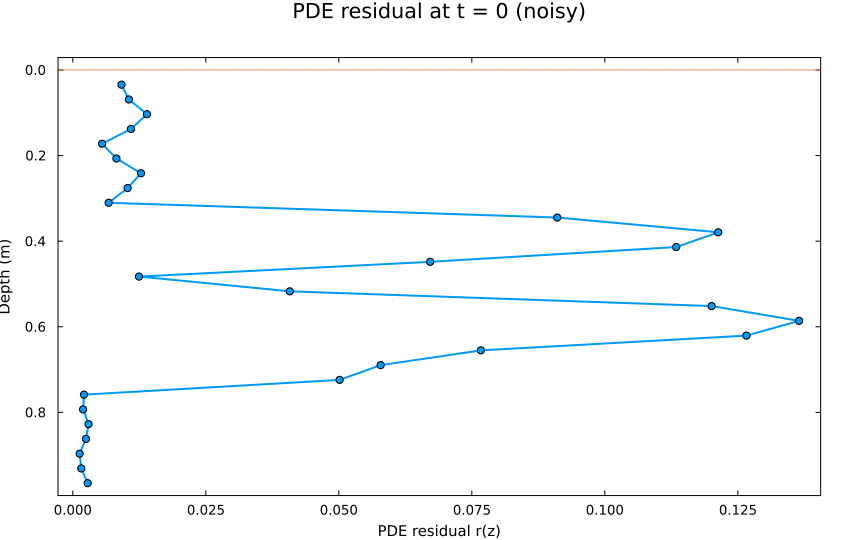}
        \caption{Residuals distribution}
        \label{fig:Residual-Distribution-Case-2}
    \end{subfigure}
    \vspace{0.8cm}
    \begin{subfigure}
        \centering        \includegraphics[width=0.6\linewidth]{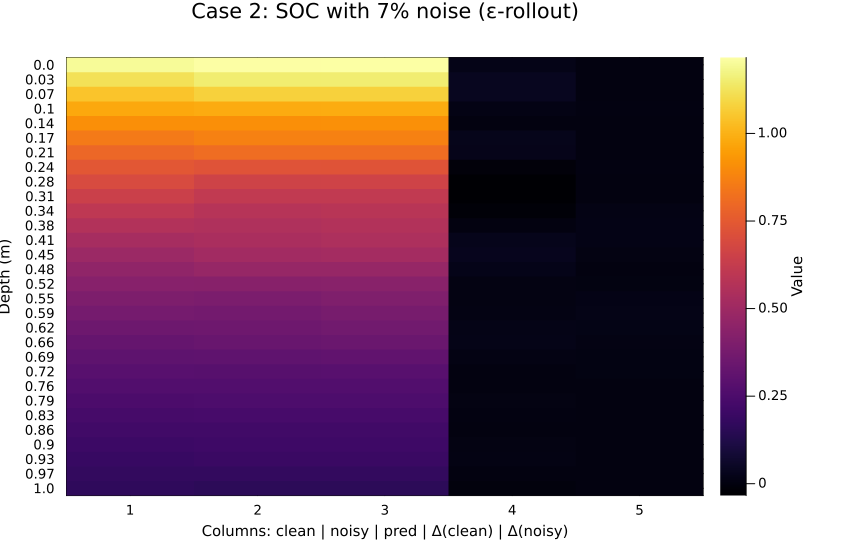}
        \caption{Heatmap of True, UDE, and Residual}
        \label{fig:Heatmap-Case-2}
    \end{subfigure}
    \caption*{Case 2 results: (a) comparison of True vs. UDE predictions, (b) residual error plot, and (c) heatmap of True SOC, UDE-predicted SOC, and residual differences.}
\end{figure}

Universal Differential Equation (UDE) results for Soil Organic Carbon (SOC) dynamics under moderate noise (7\%) at the initial profile, $t=0$ years. In this experiment, the UDE was trained on a synthetic dataset where both SOC values and soil health parameters (pH, CEC, clay content, and organic carbon inputs) were perturbed with 7\% multiplicative Gaussian noise. This setup reflects a more realistic scenario, where measurement uncertainties and natural variability are unavoidable. Panel (a) Figure~\ref{fig:True_vs_UDE-(SOC)-Case-2} compares the true SOC profile with the UDE-predicted profile across depth. Despite the noise, the UDE can capture the main structure of the SOC distribution, with only minor deviations from the ground truth. The physics-informed terms (advection and diffusion) help anchor the model to realistic depthwise behavior, while the neural component learns to smooth out noisy fluctuations. Panel (b) Figure~\ref{fig:Residual-Distribution-Case-2} shows the residual map between the prediction and ground truth. Here, residuals are more pronounced than in Case 1, with small but noticeable deviations scattered across depths. These differences illustrate the challenge of fitting noisy observations while still satisfying the governing equations. Panel (c) Figure~\ref{fig:Heatmap-Case-2} provides the heatmaps of the true SOC, the UDE prediction, and their residuals. The predicted heatmap remains closely aligned with the true SOC field, though localized patches of error can be seen in the residual map. Importantly, the UDE avoids overfitting the noise and maintains the broader SOC pattern. Overall, Case 2 demonstrates that the UDE framework is reasonably robust to moderate noise in the data. While accuracy declines slightly compared to the baseline, the model continues to reproduce the essential SOC depth profile and retains strong alignment with the underlying physical dynamics. This case highlights the balance between fitting noisy data and preserving physics-based consistency.
\subsection*{Case 3: Hyperparameter Tuning at $t=0$, High Noise (35\%)}
\begin{figure}[htbp]
    \centering
    \begin{subfigure}
        \centering        \includegraphics[width=0.6\linewidth]{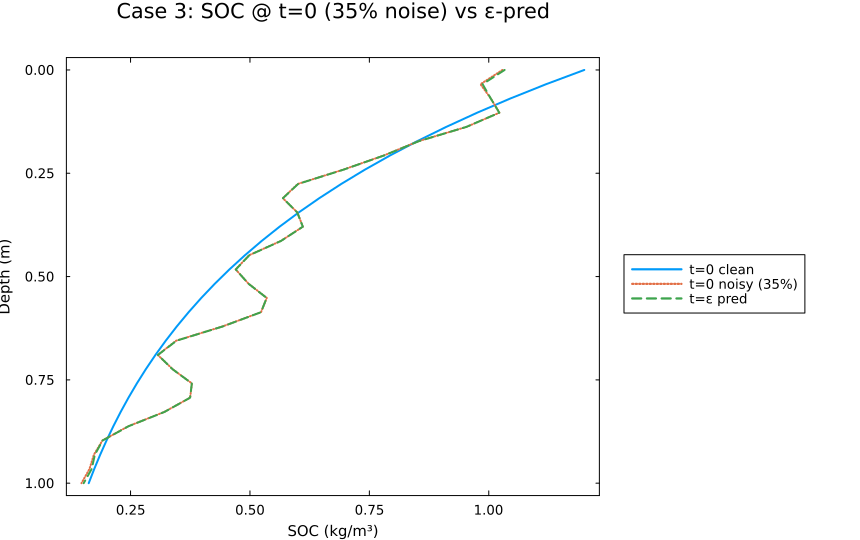}
        \caption{True vs. UDE prediction}
        \label{fig:True_vs_UDE-(SOC)-Case-3}
    \end{subfigure}
    \hfill
    \begin{subfigure}
        \centering        \includegraphics[width=0.6\linewidth]{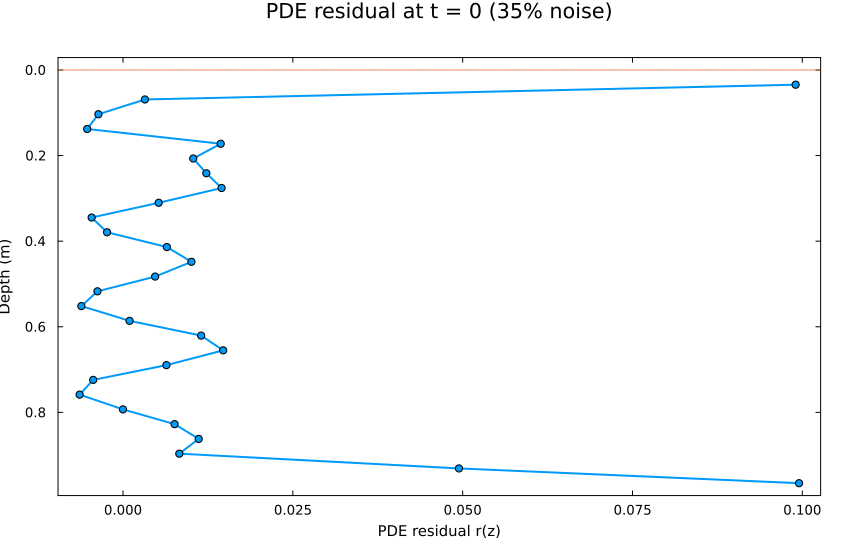}
        \caption{Residuals distribution}
        \label{fig:Residual-Distribution-Case-3}
    \end{subfigure}
    \vspace{0.8cm}
    \begin{subfigure}
        \centering        \includegraphics[width=0.6\linewidth]{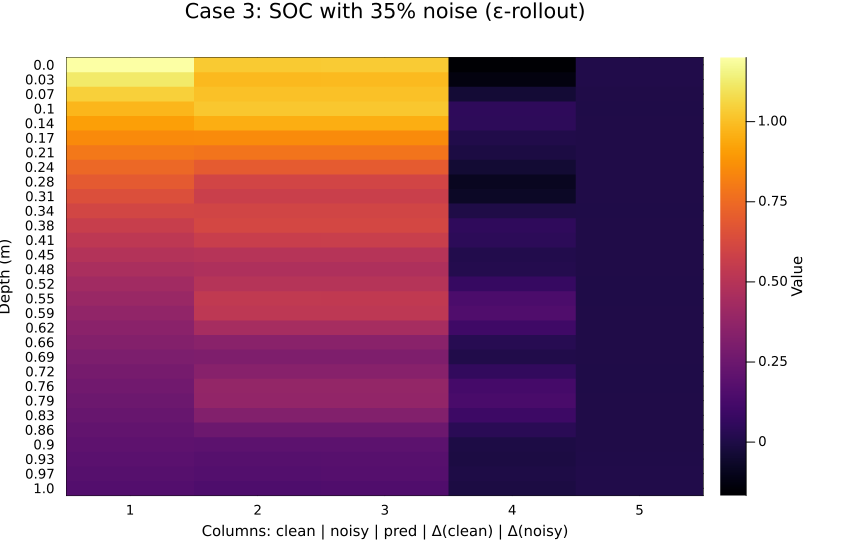}
        \caption{Heatmap of True, UDE, and Residual}
        \label{fig:Heatmap-Case-3}
    \end{subfigure}    
    \caption*{Case 3 results: (a) comparison of True vs. UDE predictions, (b) residual error plot, and (c) heatmap of True SOC, UDE-predicted SOC, and residual differences}
\end{figure}
Universal Differential Equation (UDE) results for Soil Organic Carbon (SOC) dynamics under high noise (35\%) at the initial profile, $t=0$ years. In this experiment, the UDE was trained on a synthetic dataset where both SOC values and soil health parameters (pH, CEC, clay content, and organic carbon inputs) were corrupted with 35\% multiplicative Gaussian noise. This represents a stress-test scenario, simulating conditions where measurement errors or field uncertainties are very large. Panel (a) Figure~\ref{fig:True_vs_UDE-(SOC)-Case-3} shows the comparison between the true SOC depth profile and the UDE-predicted profile. Unlike the previous cases, the predictions deviate considerably from the ground truth. While the model still follows the general trend of SOC decreasing with depth, the high noise overwhelms much of the finer structure, leading to oversmoothed or distorted profiles. The conflict between fitting highly corrupted data and obeying the governing PDE terms becomes clear in this case. Panel (b) Figure~\ref{fig:Residual-Distribution-Case-3} illustrates the residual map. Here, large and systematic deviations are visible across depths, reflecting the difficulty of extracting meaningful signals from noisy inputs. The model occasionally defaults to smoother physics-driven patterns, but these do not fully align with the noisy observations, leaving substantial errors. Panel (c) Figure~\ref{fig:Heatmap-Case-3} provides the heatmaps of the true SOC, the UDE prediction, and their residuals. Unlike in Case 2, the residual heatmap now shows significant patches of error across the entire depth range. The overlap between the true and predicted SOC fields is partial, with the UDE capturing broad trends but failing to reproduce finer details hidden by noise. Case 3 highlights the limitations of the UDE approach under severe data corruption. The model remains physically consistent but struggles to reconcile high levels of noise, leading to reduced accuracy and biased predictions. This case underscores the need for noise-aware training strategies or robust loss functions when working with highly uncertain SOC measurements.

\subsection*{Case 4: Hyperparameter Tuning at $t=50$ (Terminal SOC), No Noise (Baseline)}
\begin{figure}[htbp]
    \centering
    \begin{subfigure}
        \centering        \includegraphics[width=0.6\linewidth]{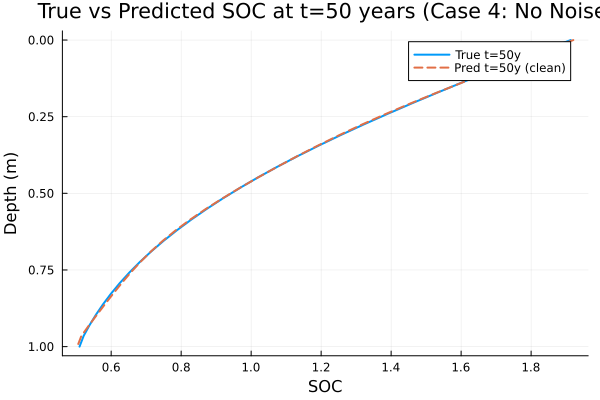}
        \caption{True vs. UDE prediction}
        \label{fig:True_vs_UDE-(SOC)-Case-4}
    \end{subfigure}
    \hfill
    \begin{subfigure}
        \centering        \includegraphics[width=0.6\linewidth]{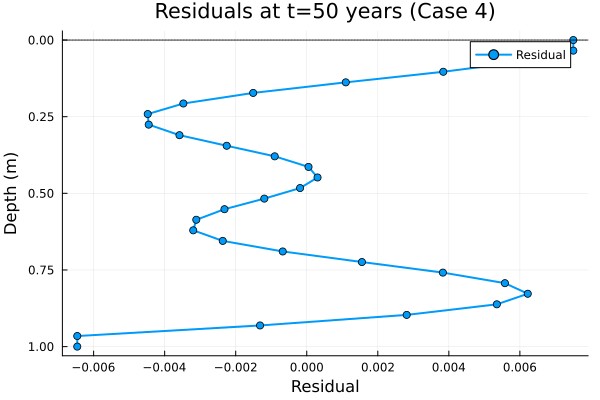}
        \caption{Residuals distribution}
        \label{fig:Residual-Distribution-Case-4}
    \end{subfigure}
    \vspace{0.8cm}
    \begin{subfigure}
        \centering        \includegraphics[width=0.6\linewidth]{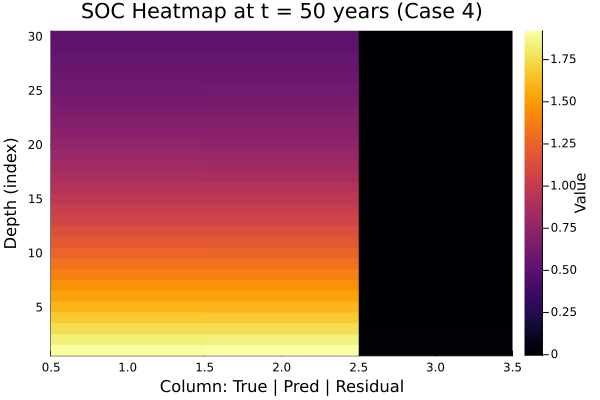}
        \caption{Heatmap of True, UDE, and Residual}
        \label{fig:Heatmap-Case-4}
    \end{subfigure}
    \caption*{Case 4 results: (a) comparison of True vs. UDE predictions, (b) residual error plot, and (c) heatmap of True SOC, UDE-predicted SOC, and residual differences.}
    \label{fig:case1_results}
\end{figure}
Universal Differential Equation (UDE) results for Soil Organic Carbon (SOC) dynamics under baseline, noise-free conditions at the terminal profile, $t=50$ years. In this case, the UDE was trained to match the final SOC distribution after 50 years of modeled evolution, using synthetic data generated from the advection–diffusion–reaction framework with soil health parameters (pH, CEC, clay content, and organic carbon inputs). Because no noise was added, this setup serves as an “ideal prediction” benchmark for long-term SOC forecasting. Panel (a) Figure~\ref{fig:True_vs_UDE-(SOC)-Case-4} shows the comparison between the true synthetic SOC profile and the UDE-predicted profile at $t=50$. The agreement is strong: the UDE accurately reproduces the depthwise SOC pattern after five decades of simulated dynamics. This confirms that, when provided with clean terminal data, the model can learn the integrated effects of both mechanistic soil processes and nonlinear biological interactions over extended time horizons. Panel (b) Figure~\ref{fig:Residual-Distribution-Case-4} presents the residual map, showing the differences between predicted and ground truth SOC values. The residuals are minimal, indicating that the UDE faithfully captures the temporal evolution and remains consistent with the governing PDE constraints. Panel (c) Figure~\ref{fig:Heatmap-Case-4} provides the heatmaps of the true SOC, the UDE prediction, and their residuals. The near-perfect overlap of the true and predicted heatmaps reinforces the accuracy of the model, while the residual heatmap stays close to zero throughout the soil depth domain. Case 4 demonstrates that the UDE framework is capable of accurately inferring long-term SOC dynamics under ideal, noise-free conditions. This result establishes a benchmark for assessing the model’s robustness in more challenging scenarios, where terminal SOC data are corrupted by measurement noise (Cases 5 and 6).

\subsection*{Case 5: Hyperparameter Tuning at t = 50 years, Moderate Noise (7\%)}
\begin{figure}[htbp]
    \centering
    \begin{subfigure}
        \centering        \includegraphics[width=0.6\linewidth]{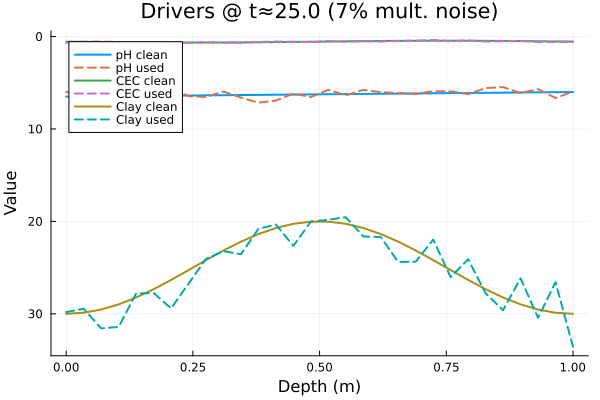}
        \caption{Drivers @ t=25.0 (7\% Multiplicative - Spatially correlated Noise)}
        \label{fig:Drivers-Soil-Parameters-Case-5}
    \end{subfigure}
    \hfill
    \begin{subfigure}
        \centering        \includegraphics[width=0.6\linewidth]{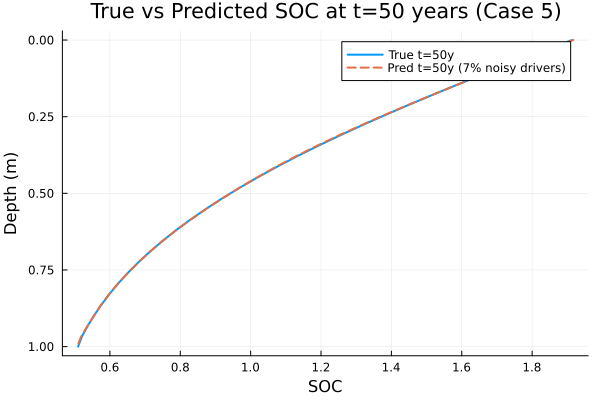}
        \caption{True vs UDE Prediction}
        \label{fig:True_vs_UDE-(SOC)-Case-5}
    \end{subfigure}
    \vspace{0.8cm}
    \begin{subfigure}
        \centering        \includegraphics[width=0.6\linewidth]{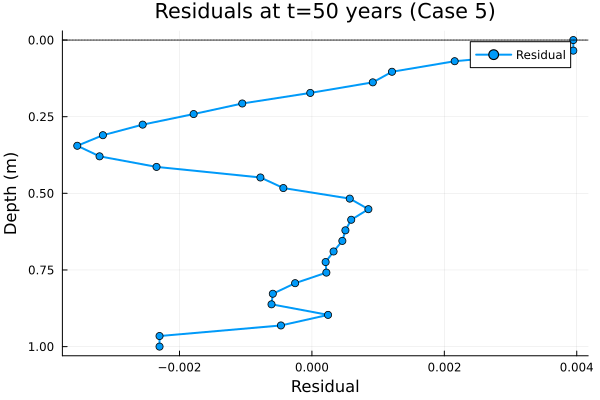}
        \caption{Residuals Distribution}       \label{fig:Residual-Distribution-Case-5}
    \end{subfigure}    
    \caption*{Case 5 results: (a) Driver (35\% Multiplicative-Spatial Noise) for soil health parameters (b) comparison of True vs. UDE predictions, and (c) residual error plot}
    \label{fig:case1_results}
\end{figure}
\vspace{0.8cm}
\begin{figure}
    \centering
    \includegraphics[width=0.6\linewidth]{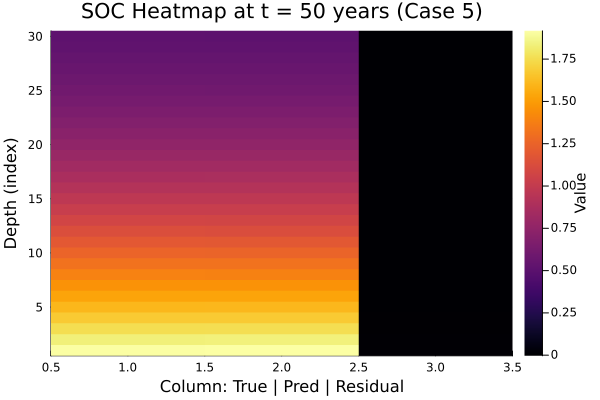}
    \caption{Heatmap of True, UDE, and Residual}
   \label{fig:Heatmap-Case-5}
\end{figure}
Universal Differential Equation (UDE) results for Soil Organic Carbon (SOC) dynamics under moderate noise (7\%) at the terminal profile, $t=50$ years. In this experiment, the UDE was trained on a synthetic dataset where both SOC values and soil health parameters were perturbed with 7\% multiplicative Gaussian noise. This reflects a realistic long-term forecasting scenario where field measurements are subject to moderate uncertainty. Panel (a) Figure~\ref{fig:Drivers-Soil-Parameters-Case-5} shows the soil drivers (pH, CEC, and clay content) at the mid-point of the simulation (t=25.0 years) under 7\% multiplicative noise. The perturbed drivers exhibit realistic fluctuations while preserving their expected depthwise trends. Panel (b) Figure~\ref{fig:True_vs_UDE-(SOC)-Case-5} compares the true terminal SOC profile with the UDE prediction. The model captures the overall depthwise pattern, though small deviations appear due to noise.  Panel (c) Figure~\ref{fig:Residual-Distribution-Case-5} shows the residual map, where errors are more noticeable than in Case 4 but remain relatively modest. Panel (d) Figure~\ref{fig:Heatmap-Case-5} presents the heatmaps of true SOC, UDE predictions, and residuals. While localized patches of error emerge, the broader SOC distribution is preserved. Case 5 demonstrates that the UDE framework is resilient to moderate noise in terminal data. Accuracy declines slightly compared to the noise-free benchmark, but the model continues to recover the main SOC trends while maintaining physics-based consistency.

\subsection*{Case 6: Hyperparameter Tuning at t = 50 years, High Noise (35\%)}
\begin{figure}[htbp]
    \centering
    \begin{subfigure}
        \centering        \includegraphics[width=0.6\linewidth]{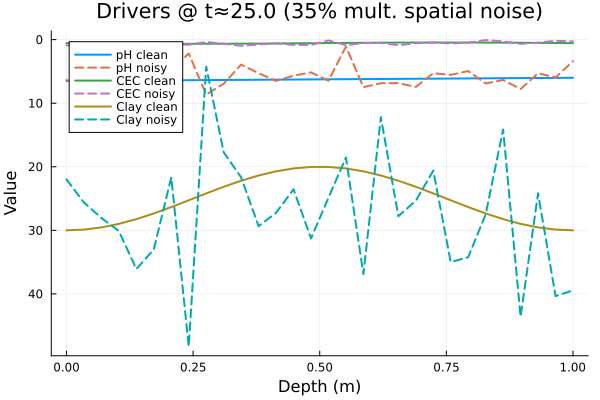}
        \caption{Drivers (35\% Multiplicative - Spatially correlated Noise) for Soil Parameters}
        \label{fig:Drivers-Soil-Parameters-Case-6}
        \vspace{2em}
    \end{subfigure}
    \hfill
    \vspace{0.2cm}    
    \begin{subfigure}
        \centering        \includegraphics[width=0.6\linewidth]{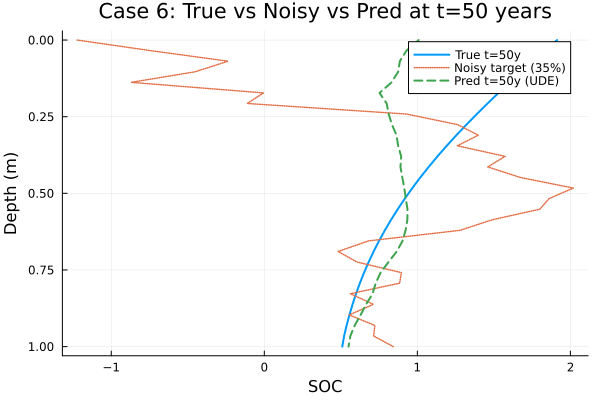}
        \caption{True vs UDE (SOC)}
        \label{fig:True_vs_UDE-(SOC)-Case-6}
    \end{subfigure}
    \vspace{0.3cm}
    \begin{subfigure}
        \centering        \includegraphics[width=0.6\linewidth]{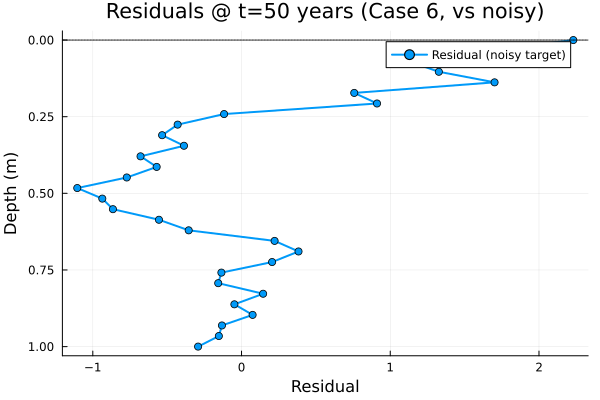}
        \caption{Residulas Distribution}
        \label{fig:Residual-Distribution-Case-6}
    \end{subfigure}    
    \caption*{Case 6 results: (a) drivers @ t=25.0 (35\% multiplicative Gaussian - spatially correlated noise (b) comparison of True vs. UDE predictions, (c) residual error plot}
    \label{fig:case1_results}
\end{figure}
\vspace{0.2cm}
\begin{figure}
    \centering
    \includegraphics[width=0.6\linewidth]{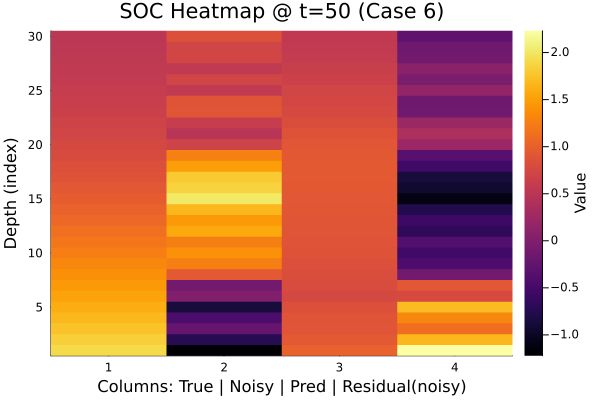}
    \caption{Heatmap of True, UDE, and Residual}
    \label{fig:Heatmap-Case-6}
\end{figure}
Universal Differential Equation (UDE) results for Soil Organic Carbon (SOC) under high noise (35\%) at the terminal profile, $t=50$ years. In this stress-test, both the terminal SOC values and soil health inputs (pH, CEC, clay content, organic carbon inputs) were corrupted with 35\% multiplicative Gaussian noise, spatially correlated across depth. This scenario mimics severely uncertain measurements and challenging field conditions. Panel (a) Figure~\ref{fig:Drivers-Soil-Parameters-Case-6} shows the soil drivers at t=25.0 years with 35\% multiplicative, spatially correlated noise. The drivers show significant fluctuations, with noticeable deviations from their clean trends, reflecting the strong perturbations imposed by high noise. Panel (b) Figure~\ref{fig:True_vs_UDE-(SOC)-Case-6} compares the true $t=50$ SOC profile with the UDE prediction across depth. Unlike the moderate-noise case, the prediction deviates substantially from the ground truth: broad trends (e.g., monotonic depth wise decline) may be partially recovered, but finer structure is lost or distorted. The model often defaults to smoother, physics-driven behaviour as it struggles to reconcile heavily corrupted targets with the governing PDE. Panel (c) Figure~\ref{fig:Residual-Distribution-Case-6} shows the residual map (prediction minus ground truth). Residuals are large and systematic over wide depth intervals, indicating that noise overwhelms the signal. The error pattern reflects the model’s trade-off: preserving PDE consistency while resisting the temptation to fit spurious fluctuations. Panel (d) Figure~\ref{fig:Heatmap-Case-6} presents heatmaps of true SOC, the UDE prediction, and residuals. The overlap between true and predicted fields is only partial; error “bands” persist throughout the profile. While the UDE maintains physically plausible smoothness, the residual heatmap reveals extensive regions of mismatch. Case 6 exposes the limits of standard UDE training under severe terminal-data noise. Predictions become biased and over-smoothed, and accuracy degrades markedly. These results underscore the need for noise-aware strategies (e.g., robust loss functions, uncertainty-weighted data terms, denoising priors) when terminal SOC observations and inputs carry high uncertainty.

\section{Discussion and Conclusions}
In this work, we developed and tested a Universal Differential Equation (UDE) framework to model soil organic carbon (SOC) dynamics across depth and time, systematically exploring six cases ranging from noise-free baseline data to high-noise stress tests. By progressively introducing noise into drivers and target SOC values, we assessed the robustness of UDEs in reconstructing SOC evolution and predicting long-term outcomes.

For Case 1 (Clean baseline, t=0 years), the UDE successfully reproduced the synthetic SOC distribution using noise-free drivers and terminal data. This case confirmed that the architecture, loss formulation, and optimization strategy were well-suited for the problem, yielding near-perfect fits and stable residuals.

In Case 2 (Clean drivers with collocation points), we extended training to include intermediate times, enforcing PDE consistency beyond terminal conditions. This led to smoother solutions and improved generalization, particularly in capturing transport and diffusion effects. The inclusion of collocation points demonstrated the utility of physics-based constraints for stabilizing learning even in data-sparse regimes.

Case 3 (High noise, 35\%) represented a worst-case scenario where both drivers and final SOC values were heavily corrupted. As expected, the UDE struggled in this regime, with predictions biased toward satisfying the PDE rather than reproducing noisy observations. Nevertheless, the model maintained physically plausible trajectories, suggesting that UDEs can filter extreme noise to some degree, though at the expense of accuracy. This highlights the limitation of standard loss formulations under very high uncertainty, underscoring the need for robust loss functions or explicit noise modelling in future work.

In Case 4 (No Noise, t=50 years), where the model was trained to recover long-term SOC profiles from perfect terminal conditions, the UDE excelled. Despite the inverse nature of the problem (learning dynamics from end states), the model reconstructed depth-wise SOC distributions with high fidelity. This case illustrated the power of UDEs in capturing long-term integrated behavior when precise boundary or terminal data are available.

Case 5 (Medium noise, 7\%) introduced realistic uncertainties in both drivers and SOC targets. The model retained good predictive performance, with moderate residuals, though slightly biased by the noisy training signals. This case best reflects real-world SOC monitoring conditions, where field data often contain moderate but unavoidable measurement errors. The results demonstrate that UDEs can still yield useful forecasts under such conditions, balancing physics-informed dynamics with noisy observations.

Finally, Case 6 (High noise, 35\%, t=50 years) served as the ultimate stress test. Similar to Case 3, the model prioritized enforcing PDE structure over fitting corrupted SOC values. Predictions were smooth but diverged from the noisy data, confirming that without noise-aware loss terms or denoising strategies, UDEs have limited reliability under extreme uncertainty. This outcome is consistent with expectations and sets the stage for future research on robust training methodologies.

\subsection{\textbf{Hyperparameters for six different cases}}
To systematically evaluate the robustness of the proposed SOC–UDE framework, we designed six experimental cases that progressively increase the complexity and noise levels in the synthetic dataset. Each case uses the same baseline soil drivers—pH, cation exchange capacity (CEC), and clay content—defined across the spatial domain ($z \in [0,1]$) and temporal horizon (t = 50 years). The difference between the cases lies in how measurement noise is injected into the drivers and terminal SOC data, thereby simulating the uncertainty and variability typically present in real soil datasets. For each case, we systematically tune key hyperparameters—including network depth and width (hidden units), activation functions, learning rate schedules, optimization algorithms (Adam, BFGS), and loss weights that balance terminal SOC fitting, collocation residuals, and mild weight decay. This grid-style search helps to identify robust model configurations capable of capturing SOC dynamics under different levels of uncertainty.

The following Hyperparameter tables summarize the chosen hyperparameters and their search ranges for each experimental case.

\begin{table}[H]
\caption{Hyperparameters for Case 1 – Clean synthetic baseline (no noise)}
\begin{flushleft}
\begin{tabular}{lll}
\toprule
Hyperparameter & Values & Search range \\
\midrule
Spatial grid (Nz)    & 30                       & 20--40 \\
Time span (tspan)    & (0, 50 years)            & Fixed \\
Activation Function  & tanh, GELU               & \{tanh, GELU\} \\
Optimizers           & Adam (3e-3), BFGS        & Adam, BFGS \\
Learning Rate (LR)   & 3e-3 (Adam), 1e-2 (BFGS) & 1e-3--1e-2 \\
Hidden units (H1,H2) & (32,16)                  & \{16,32,64\} $\times$ \{8,16,32\} \\
Epochs (iters)       & 200 (Adam), 400 (BFGS)   & 100--1000 \\
Loss weights         & $\lambda_{term}=1$, $\lambda_{coll}=1$, $\lambda_{wd}=10^{-4}$ & Fixed \\
\bottomrule
\end{tabular}
\end{flushleft}
\label{tab:case1}
\end{table}

\begin{table}[H]
\caption{Hyperparameters for Case 2 – Moderate noise (7\%) in drivers \& SOC}
\begin{flushleft}
\begin{tabular}{lll}
\toprule
Hyperparameter & Values & Search range \\
\midrule
Driver noise         & 7\% multiplicative, spatial AR-1 & 5--10\% \\
Terminal SOC noise   & 7\% Gaussian                    & 5--10\% \\
Activation Function  & tanh, GELU                      & \{tanh, GELU\} \\
Optimizers           & Adam (3e-3), BFGS               & Adam, BFGS \\
Learning Rate (LR)   & 3e-3                            & 1e-3--1e-2 \\
Hidden units (H1,H2) & (32,32)                         & \{32--64\} $\times$ \{16--64\} \\
Epochs (iters)       & 200 (Adam), 400 (BFGS)          & 200--800 \\
Loss weights         & $\lambda_{term}=1.5$, $\lambda_{coll}=1$, $\lambda_{wd}=10^{-3}$ & $\lambda_{term} \in \{1-2\}$ \\
\bottomrule
\end{tabular}
\end{flushleft}
\label{tab:case5}
\end{table}

\begin{table}[H]
\caption{Hyperparameters for Case 3 – High Noise (35\%) in drivers \& SOC}
\begin{flushleft}
\begin{tabular}{lll}
\toprule
Hyperparameter & Values & Search range \\
\midrule
Driver noise         & 35\% multiplicative, spatially correlated & 30--40\% \\
Terminal SOC noise   & 35\% Gaussian                             & 30--40\% \\
Activation Function  & tanh, GELU                                & \{tanh, GELU\} \\
Optimizers           & Adam (3e-3), BFGS                         & Adam, BFGS \\
Learning Rate (LR)   & 3e-3                                      & 1e-3--5e-3 \\
Hidden units (H1,H2) & (32,32)                                   & \{32--64\} $\times$ \{16--64\} \\
Epochs (iters)       & 200 (Adam), 400 (BFGS)                    & 200--800 \\
Loss weights         & $\lambda_{term}=1$, $\lambda_{coll}=1$, $\lambda_{wd}=10^{-3}$ & $\lambda_{wd} \in \{10^{-4},10^{-3},10^{-2}\}$ \\
\bottomrule
\end{tabular}
\end{flushleft}
\label{tab:case3_highnoise}
\end{table}

\begin{table}[H]
\caption{Hyperparameters for Case 4 – No Noise, terminal SOC at $t=50$ years}
\begin{flushleft}
\begin{tabular}{lll}
\toprule
Hyperparameter & Values & Search range \\
\midrule
Driver noise         & None (clean drivers)         & Fixed \\
Terminal SOC noise   & None (clean target at $t=50$) & Fixed \\
Activation Function  & tanh, GELU                   & \{tanh, GELU\} \\
Optimizers           & Adam (3e-3), BFGS            & Adam, BFGS \\
Learning Rate (LR)   & 3e-3                         & 1e-3--5e-3 \\
Hidden units (H1,H2) & (32,16)                      & \{16--64\} $\times$ \{8--32\} \\
Epochs (iters)       & 200 (Adam), 400 (BFGS)       & 200--600 \\
Loss weights         & $\lambda_{term}=1$, $\lambda_{coll}=1$, $\lambda_{wd}=10^{-4}$ & Fixed \\
\bottomrule
\end{tabular}
\end{flushleft}
\label{tab:case4_nonoise}
\end{table}

\begin{table}[H]
\caption{Hyperparameters for Case 5 – Medium Noise (7\%), terminal SOC at $t=50$ years}
\begin{flushleft}
\begin{tabular}{lll}
\toprule
Hyperparameter & Values & Search range \\
\midrule
Driver noise         & 7\% multiplicative, spatial AR-1 & 5--10\% \\
Terminal SOC noise   & 7\% Gaussian                    & 5--10\% \\
Activation Function  & tanh, GELU                      & \{tanh, GELU\} \\
Optimizers           & Adam (3e-3), BFGS               & Adam, BFGS \\
Learning Rate (LR)   & 3e-3                            & 1e-3--1e-2 \\
Hidden units (H1,H2) & (32,32)                         & \{32--64\} $\times$ \{16--64\} \\
Epochs (iters)       & 200 (Adam), 400 (BFGS)          & 200--800 \\
Loss weights         & $\lambda_{term}=1.5$, $\lambda_{coll}=1$, $\lambda_{wd}=10^{-3}$ & $\lambda_{term} \in \{1-2\}$ \\
\bottomrule
\end{tabular}
\end{flushleft}
\label{tab:case5_mediumnoise}
\end{table}

\begin{table}[H]
\caption{Hyperparameters for Case 6 – High Noise (35\%), terminal SOC at $t=50$ years}
\begin{flushleft}
\begin{tabular}{lll}
\toprule
Hyperparameter & Values & Search range \\
\midrule
Driver noise         & 35\% multiplicative, spatial AR-1 & 30--40\% \\
Terminal SOC noise   & 35\% Gaussian                     & 30--40\% \\
Activation Function  & tanh, GELU                        & \{tanh, GELU\} \\
Optimizers           & Adam (1e-3), BFGS                 & Adam, BFGS \\
Learning Rate (LR)   & 1e-3                              & 5e-4--5e-3 \\
Hidden units (H1,H2) & (64,32)                           & \{32--128\} $\times$ \{16--64\} \\
Epochs (iters)       & 400 (Adam), 800 (BFGS)            & 400--1000 \\
Loss weights         & $\lambda_{term}=2.0$, $\lambda_{coll}=1.0$, $\lambda_{wd}=10^{-4}$ & $\lambda_{term} \in \{1.5 - 2.5\}$ \\
\bottomrule
\end{tabular}
\end{flushleft}
\label{tab:case6_highnoise}
\end{table}

\subsection{Important Results}
The Case 3 (t=0) examined training a UDE under severe multiplicative spatial noise (35\%). A 144-trial hyperparameter sweep found H1=32, H2=16, tanh, LR=0.001 with \(\lambda_{pde}=5\) and \(\lambda_{data}=0.25\) as best. The tuned model closely fits the corrupted labels (RMSE=0.0018, \(R^2\)=0.99996) but shows larger error versus the clean truth (RMSE=0.075, \(R^2\)=0.938), indicating partial overfitting to noise. PDE residuals remain nonzero, reflecting a trade-off between data fidelity and physical consistency. For practical use we recommend robust losses, explicit noise models, stronger PDE weighting, or ensembling to recover the underlying clean signal. Cross-validation and early stopping further mitigate memorization and improve generalization robustly now.

In Case 4 (t=50), the UDE model was trained on clean, noise-free synthetic drivers (pH, CEC, clay) and SOC profiles, providing a benchmark scenario. A grid search of 16 hyperparameter combinations tested different hidden layer sizes, activations (\texttt{tanh}, \texttt{GELU}), and learning rates. The best configuration (\(H_1=32\), \(H_2=16\), \texttt{GELU}, LR=\(0.003\)) achieved the lowest tuning loss (\(1.1 \times 10^{-4}\)). After retraining, the model reproduced the terminal SOC profile with very high fidelity, yielding MSE=\(1.6 \times 10^{-5}\), and \(R^2=0.9999\). This demonstrates the UDE’s ability to recover SOC dynamics almost perfectly under clean conditions.  

In the Case 5 (t=50), the UDE was trained on synthetic drivers (pH, CEC, clay) and SOC profiles perturbed with 7\% multiplicative Gaussian noise, simulating realistic measurement uncertainty. A grid search of 16 hyperparameter combinations was tested, covering hidden layer sizes, activations (\texttt{tanh}, \texttt{GELU}), and learning rates. The best configuration (\(H_1=32\), \(H_2=16\), \texttt{tanh}, LR=\(0.003\)) achieved the lowest tuning loss (\(2.2 \times 10^{-6}\)). After retraining, the model reproduced the noisy terminal SOC profile with excellent accuracy, yielding MSE=\(3.4 \times 10^{-6}\) and \(R^2=0.99998\). Despite the added uncertainty, the UDE remained robust and preserved the main SOC dynamics effectively.  

In this experiment (Case 6, t=50), the UDE was trained on synthetic drivers and SOC profiles heavily perturbed with 35\% multiplicative, spatially correlated noise. This represents an extreme stress test where both inputs and terminal SOC targets are strongly corrupted, mimicking challenging field conditions. A grid of 16 hyperparameter settings was explored, and the best-performing configuration (\(H_1=64\), \(H_2=32\), \texttt{tanh}, LR=\(0.003\)) achieved a tuning loss of \(0.99\). After retraining, the model reached a final noisy-target MSE of \(0.694\), but the $R^2$ was negative, reflecting the difficulty of fitting to highly corrupted data. When benchmarked against the clean ground truth, performance slightly improved (RMSE=\(0.44\)), yet $R^2$ remained negative. Case 6 demonstrates that under extreme noise, the UDE struggles to recover SOC dynamics, highlighting the limits of robustness under severe measurement uncertainty.  The imporant results of Case 3 to 6, are tabulated below for ready reference.
\begin{table}[H]
\centering
\caption{Comparative Summary of UDE-based SOC Modeling Results (Cases 3–6)}
\label{tab:comparative_summary}
\begin{tabular}{|c|p{4.0cm}|c|c|}
\hline
\textbf{Case} & \textbf{Best Hyperparameters} & \textbf{MSE} & \textbf{$R^2$} \\ \hline
3 (t=0, 35\% Noise) & H1=32, H2=16, ACT=tanh, LR=0.001, $\lambda_{\text{pde}}=5.0$, $\lambda_{\text{data}}=0.25$ & – & 0.938 (clean), 0.99996 (noisy) \\ \hline
4 (t=50, No Noise) & H1=32, H2=16, ACT=gelu, LR=0.003 & $1.60\times10^{-5}$ & 0.9999 \\ \hline
5 (t=50, 7\% Noise) & H1=32, H2=16, ACT=tanh, LR=0.003 & $3.39\times10^{-6}$ & 0.99998 \\ \hline
6 (t=50, 35\% Noise) & H1=64, H2=32, ACT=tanh, LR=0.003 & 0.192 (clean), 0.694 (noisy) & –0.033 (clean), –0.017 (noisy) \\ \hline
\end{tabular}
\end{table}

\subsection{\textbf{Conclusions and Challenges}}
This study explored the potential of Universal Differential Equations (UDEs) for forecasting Soil Organic Carbon (SOC) dynamics across depth and time. By embedding advection–diffusion transport within a neural framework, the model combined mechanistic realism with the flexibility of Scientific Machine Learning (SciML). Six experimental cases were evaluated, ranging from clean, noise-free baselines to severe high-noise stress tests. 

Quantitatively, the results demonstrated that UDEs excel under noise-free and moderate-noise conditions. In Case~4 (clean, $t=50$), the model achieved near-perfect fidelity ($\mathrm{MSE}=1.6 \times 10^{-5}$, $R^{2}=0.9999$), while Case~5 (7\% noise) maintained robustness ($\mathrm{MSE}=3.4 \times 10^{-6}$, $R^{2}=0.99998$). In contrast, Case~3 (35\% noise at $t=0$) revealed overfitting to corrupted inputs, reducing $R^{2}$ against clean truth to 0.94, and Case~6 (35\% noise at $t=50$) yielded negative $R^{2}$, reflecting the limitations of standard loss functions under severe uncertainty. 

Hyperparameter tuning highlighted several best practices. Two-layer architectures with modest hidden sizes (H1=32–64, H2=16–32) balanced expressivity and stability. Smooth activations such as \texttt{tanh} and GELU outperformed ReLU, avoiding solver instabilities. Training benefited from combining Adam for initial exploration with BFGS for fine convergence. Importantly, results showed that complexity beyond shallow networks offered little additional benefit under synthetic settings.

Overall, this work confirms that UDEs provide a scalable, physics-informed alternative for SOC modelling, particularly resilient to moderate measurement noise. At the same time, their fragility under high-noise conditions emphasizes the need for noise-aware loss functions, probabilistic formulations, and stronger integration of microbial dynamics. Addressing these challenges will be essential for advancing from synthetic benchmarks to reliable field-scale SOC prediction, enabling applications in carbon monitoring, sustainable soil management, and climate policy.

\textbf{Challenges:} The Universal Differential Equation (UDE) framework offers a promising way to combine soil process physics with the flexibility of machine learning. Yet, our current setup embodies strong simplifications that introduce important challenges. Training on synthetic data, for instance, avoids the messiness of real-world measurements but risks overfitting to idealized patterns. In practice, soil carbon observations are noisy, sparse, and context-specific; a model trained only on clean synthetic data may perform poorly when confronted with real variability.

Similarly, the use of sealed, no-flux boundary conditions simplifies the mathematics but ignores real carbon exchanges at the soil surface and deeper layers. This can trap or exclude flows that matter for long-term SOC dynamics. The choice of constant advection–diffusion coefficients further overlooks the inherent heterogeneity of soils: texture, moisture, and structure vary with depth and location, meaning transport rates are rarely uniform. These assumptions, while convenient, risk biasing predictions in ways that remain hidden until tested against field data.

Finally, using simple neural networks to approximate complex source and sink processes can limit expressivity. While small networks are easier to train, they may smooth over sharp nonlinear responses—such as rapid decomposition after rainfall—while larger networks risk overfitting when data are scarce.

The scientific literature on soil carbon modeling and physics-informed ML highlights these dangers, emphasizing the need for robustness to noise, flexibility to capture heterogeneity, and caution with model complexity. Without these safeguards, UDE-based forecasts may yield misleading estimates, undermining trust in carbon markets and climate policy applications.

\subsection{Future Research Directions in SOC UDE Modeling}
The SOC UDE framework presented here opens several promising avenues for further research. One important direction is the incorporation of probabilistic drivers, which would allow uncertainty in climatic and soil parameters to be explicitly modeled, thereby capturing the stochastic nature of SOC dynamics. Another extension involves the application of symbolic regression techniques on the learned terms to enhance physical interpretability, bridging the gap between data-driven models and established soil science theory. Additionally, scaling the framework to spatially distributed SOC–climate interactions would make it possible to quantify regional or global patterns of carbon sequestration under varying land-use and climate scenarios. 

Beyond these methodological advances, a critical frontier lies in better integrating microbial dynamics into SOC UDE models. Although microbial process modeling has seen substantial progress, there remains a significant divide between what microbial ecologists can measure and what biogeochemists can feasibly represent in models. Closing this gap requires UDEs that not only capture microbial processes such as decomposition, stabilization, and priming effects, but also remain robust enough to be validated against diverse field datasets. By embedding microbial mechanisms into a physics-informed learning framework, SOC UDE models could more faithfully represent the interplay between biotic and abiotic factors that govern carbon cycling. Together, these directions will strengthen the predictive capacity of SOC UDEs, enabling them to serve as valuable tools for scientific inquiry, sustainable soil management, and evidence-based policy design in the context of climate change mitigation.

\clearpage
\bibliographystyle{unsrtnat}
\bibliography{references}  

\begin{thebibliography}{76}
\providecommand{\natexlab}[1]{#1}
\providecommand{\url}[1]{\texttt{#1}}
\expandafter\ifx\csname urlstyle\endcsname\relax
  \providecommand{\doi}[1]{doi: #1}\else
  \providecommand{\doi}{doi: \begingroup \urlstyle{rm}\Url}\fi

\bibitem[Bhattacharyya et~al.(2011)Bhattacharyya, Pal, Chandran, Ray, Mandal, Wani, and Sahrawat]{bhattacharyya2011carbon}
T~Bhattacharyya, DK~Pal, P~Chandran, SK~Ray, C~Mandal, SP~Wani, and KL~Sahrawat.
\newblock Carbon status of indian soils: an overview.
\newblock 2011.

\bibitem[Gurmu(2019)]{gurmu2019soil}
Gebreyes Gurmu.
\newblock Soil organic matter and its role in soil health and crop productivity improvement.
\newblock \emph{Forest Ecology and Management}, 7\penalty0 (7):\penalty0 475--483, 2019.

\bibitem[Jha et~al.(2023)Jha, Bonetti, Smith, Souza, and Calabrese]{Jha2023}
Achla Jha, Sara Bonetti, A.~Peyton Smith, Rodolfo Souza, and Salvatore Calabrese.
\newblock Linking soil structure, hydraulic properties, and organic carbon dynamics: A holistic framework to study the impact of climate change and land management, 1 2023.

\bibitem[Xu et~al.(2021)Xu, Li, Kuyper, Xu, Li, and Zhang]{Xu2021}
Meng Xu, Xiaoliang Li, Thomas~W. Kuyper, Ming Xu, Xiaolin Li, and Junling Zhang.
\newblock High microbial diversity stabilizes the responses of soil organic carbon decomposition to warming in the subsoil on the tibetan plateau.
\newblock \emph{Global Change Biology}, 27:\penalty0 2061--2075, 5 2021.
\newblock ISSN 13652486.
\newblock \doi{10.1111/gcb.15553}.

\bibitem[Ballantyne and Billings(2018)]{Ballantyne2018}
Ford Ballantyne and Sharon~A. Billings.
\newblock Model formulation of microbial co2 production and efficiency can significantly influence short and long term soil c projections, 6 2018.
\newblock ISSN 17517370.

\bibitem[Monti et~al.(2024)Monti, Diele, Lacitignola, and Marangi]{Monti2024}
Angela Monti, Fasma Diele, Deborah Lacitignola, and Carmela Marangi.
\newblock Patterns in soil organic carbon dynamics: integrating microbial activity, chemotaxis and data-driven approaches.
\newblock 7 2024.
\newblock URL \url{http://arxiv.org/abs/2407.20625}.

\bibitem[Ortner et~al.(2022)Ortner, Seidel, Semella, Udelhoven, Vohland, and Thiele-Bruhn]{ortner2022content}
Malte Ortner, Michael Seidel, Sebastian Semella, Thomas Udelhoven, Michael Vohland, and S{\"o}ren Thiele-Bruhn.
\newblock Content of soil organic carbon and labile fractions depend on local combinations of mineral-phase characteristics.
\newblock \emph{Soil}, 8\penalty0 (1):\penalty0 113--131, 2022.

\bibitem[John et~al.(2020)John, Abraham~Isong, Michael~Kebonye, Okon~Ayito, Chapman~Agyeman, and Marcus~Afu]{john2020using}
Kingsley John, Isong Abraham~Isong, Ndiye Michael~Kebonye, Esther Okon~Ayito, Prince Chapman~Agyeman, and Sunday Marcus~Afu.
\newblock Using machine learning algorithms to estimate soil organic carbon variability with environmental variables and soil nutrient indicators in an alluvial soil.
\newblock \emph{Land}, 9\penalty0 (12):\penalty0 487, 2020.

\bibitem[Singh et~al.(2010)Singh, Bardgett, Smith, and Reay]{Singh2010}
Brajesh~K. Singh, Richard~D. Bardgett, Pete Smith, and Dave~S. Reay.
\newblock Microorganisms and climate change: Terrestrial feedbacks and mitigation options, 11 2010.
\newblock ISSN 17401526.

\bibitem[Ding et~al.(2025)Ding, Liu, Grunwald, Smith, Ciais, Wang, Wadoux, Ferreira, Karunaratne, Shurpali, et~al.]{ding2025advancing}
Zijuan Ding, Ke~Liu, Sabine Grunwald, Pete Smith, Philippe Ciais, Bin Wang, Alexandre MJ-C Wadoux, Carla Ferreira, Senani Karunaratne, Narasinha Shurpali, et~al.
\newblock Advancing soil organic carbon prediction: A comprehensive review of technologies, ai, process-based and hybrid modelling approaches.
\newblock \emph{Advanced Science}, page e04152, 2025.

\bibitem[Li et~al.(2024)Li, Cui, Wu, McLaren, Xia, Pandey, Liu, Wang, Xu, Song, Dalal, and Dang]{Li2024}
Tong Li, Lizhen Cui, Yu~Wu, Timothy~I. McLaren, Anquan Xia, Rajiv Pandey, Hongdou Liu, Weijin Wang, Zhihong Xu, Xiufang Song, Ram~C. Dalal, and Yash~P. Dang.
\newblock Soil organic carbon estimation via remote sensing and machine learning techniques: Global topic modeling and research trend exploration.
\newblock \emph{Remote Sensing}, 16, 9 2024.
\newblock ISSN 20724292.
\newblock \doi{10.3390/rs16173168}.

\bibitem[Ahrens et~al.(2015)Ahrens, Braakhekke, Guggenberger, Schrumpf, and Reichstein]{Ahrens2015}
Bernhard Ahrens, Maarten~C. Braakhekke, Georg Guggenberger, Marion Schrumpf, and Markus Reichstein.
\newblock Contribution of sorption, doc transport and microbial interactions to the 14c age of a soil organic carbon profile: Insights from a calibrated process model.
\newblock \emph{Soil Biology and Biochemistry}, 88:\penalty0 390--402, 9 2015.
\newblock ISSN 00380717.
\newblock \doi{10.1016/j.soilbio.2015.06.008}.

\bibitem[Wang et~al.(2024)Wang, Abramowitz, Wang, Pitman, and Viscarra~Rossel]{wang2024ensemble}
Lingfei Wang, Gab Abramowitz, Ying-Ping Wang, Andy Pitman, and Raphael~A Viscarra~Rossel.
\newblock An ensemble estimate of australian soil organic carbon using machine learning and process-based modelling.
\newblock \emph{Soil}, 10\penalty0 (2):\penalty0 619--636, 2024.

\bibitem[Neofytou et~al.(2024)Neofytou, Neophytides, Eliades, Papoutsa, Tzouvaras, and Hadjimitsis]{neofytou2024review}
Eleni Neofytou, Stelios~P Neophytides, Marinos Eliades, Christiana Papoutsa, Marios Tzouvaras, and Diofantos~G Hadjimitsis.
\newblock A review of soil organic carbon (soc) prediction techniques in agricultural lands using remote sensing.
\newblock In \emph{IGARSS 2024-2024 IEEE International Geoscience and Remote Sensing Symposium}, pages 1273--1279. IEEE, 2024.

\bibitem[Kakhani et~al.(2024)Kakhani, Rangzan, Jamali, Attarchi, Alavipanah, Mommert, Tziolas, and Scholten]{Kakhani2024}
Nafiseh Kakhani, Moien Rangzan, Ali Jamali, Sara Attarchi, Seyed~Kazem Alavipanah, Michael Mommert, Nikolaos Tziolas, and Thomas Scholten.
\newblock Ssl-soilnet: A hybrid transformer-based framework with self-supervised learning for large-scale soil organic carbon prediction.
\newblock 8 2024.
\newblock URL \url{http://arxiv.org/abs/2308.03586}.

\bibitem[Berardi et~al.(2024)Berardi, Hartman, Brzostek, Bernacchi, DeLucia, von Haden, Kantola, Moore, Yang, Hudiburg, and Parton]{Berardi2024}
Danielle~M. Berardi, Melannie~D. Hartman, Edward~R. Brzostek, Carl~J. Bernacchi, Evan~H. DeLucia, Adam~C. von Haden, Ilsa Kantola, Caitlin~E. Moore, Wendy~H. Yang, Tara~W. Hudiburg, and William~J. Parton.
\newblock Microbial-explicit processes and refined perennial plant traits improve modeled ecosystem carbon dynamics.
\newblock \emph{Geoderma}, 443, 3 2024.
\newblock ISSN 00167061.
\newblock \doi{10.1016/j.geoderma.2024.116851}.

\bibitem[Luo et~al.(2017)Luo, Feng, Luo, Baldock, and Wang]{luo2017soil}
Zhongkui Luo, Wenting Feng, Yiqi Luo, Jeff Baldock, and Enli Wang.
\newblock Soil organic carbon dynamics jointly controlled by climate, carbon inputs, soil properties and soil carbon fractions.
\newblock \emph{Global change biology}, 23\penalty0 (10):\penalty0 4430--4439, 2017.

\bibitem[Bursa{\'c} et~al.(2022)Bursa{\'c}, Kova{\v{c}}evi{\'c}, and Bajat]{bursac2022instance}
Petar Bursa{\'c}, Milo{\v{s}} Kova{\v{c}}evi{\'c}, and Branislav Bajat.
\newblock Instance-based transfer learning for soil organic carbon estimation.
\newblock \emph{Frontiers in Environmental Science}, 10:\penalty0 1003918, 2022.

\bibitem[Ermolieva et~al.(2024{\natexlab{a}})Ermolieva, Havlik, Lessa-Derci-Augustynczik, Frank, Balkovic, Skalsky, Deppermann, Nakhavali, Komendantova, Kahil, et~al.]{ermolieva2024tracking}
Tatiana Ermolieva, Petr Havlik, Andrey Lessa-Derci-Augustynczik, Stefan Frank, Juraj Balkovic, Rastislav Skalsky, Andre Deppermann, Mahdi Nakhavali, Nadejda Komendantova, Taher Kahil, et~al.
\newblock Tracking the dynamics and uncertainties of soil organic carbon in agricultural soils based on a novel robust meta-model framework using multisource data.
\newblock \emph{Sustainability}, 16\penalty0 (16):\penalty0 6849, 2024{\natexlab{a}}.

\bibitem[Pavlovic et~al.(2024{\natexlab{a}})Pavlovic, Ilic, Ralevic, Antonic, Raffa, Bandecchi, and Culibrk]{pavlovic2024deep}
Marko Pavlovic, Slobodan Ilic, Neoboj{\v{s}}a Ralevic, Nenad Antonic, Dylan~Warren Raffa, Michele Bandecchi, and Dubravko Culibrk.
\newblock A deep learning approach to estimate soil organic carbon from remote sensing.
\newblock \emph{Remote Sensing}, 16\penalty0 (4):\penalty0 655, 2024{\natexlab{a}}.

\bibitem[Ku{\'s}mierz et~al.(2023)Ku{\'s}mierz, Skowro{\'n}ska, Tkaczyk, Lipi{\'n}ski, and Mielniczuk]{kusmierz2023soil}
Sebastian Ku{\'s}mierz, Monika Skowro{\'n}ska, Przemys{\l}aw Tkaczyk, Wojciech Lipi{\'n}ski, and Jacek Mielniczuk.
\newblock Soil organic carbon and mineral nitrogen contents in soils as affected by their ph, texture and fertilization.
\newblock \emph{Agronomy}, 13\penalty0 (1):\penalty0 267, 2023.

\bibitem[M{\"a}kip{\"a}{\"a} et~al.(2024)M{\"a}kip{\"a}{\"a}, Menichetti, Mart{\'\i}nez-Garc{\'\i}a, T{\"o}rm{\"a}nen, and Lehtonen]{makipaa2024organic}
Raisa M{\"a}kip{\"a}{\"a}, Lorenzo Menichetti, Eduardo Mart{\'\i}nez-Garc{\'\i}a, Tiina T{\"o}rm{\"a}nen, and Aleksi Lehtonen.
\newblock Is the organic carbon-to-clay ratio a reliable indicator of soil health?
\newblock \emph{Geoderma}, 444:\penalty0 116862, 2024.

\bibitem[Solly et~al.(2020)Solly, Weber, Zimmermann, Walthert, Hagedorn, and Schmidt]{solly2020critical}
Emily~F Solly, Valentino Weber, Stephan Zimmermann, Lorenz Walthert, Frank Hagedorn, and Michael~WI Schmidt.
\newblock A critical evaluation of the relationship between the effective cation exchange capacity and soil organic carbon content in swiss forest soils.
\newblock \emph{Frontiers in Forests and Global Change}, 3:\penalty0 98, 2020.

\bibitem[Zech et~al.(2024)Zech, Prechtel, and Ray]{Zech2024}
Simon Zech, Alexander Prechtel, and Nadja Ray.
\newblock Coupling scales in process-based soil organic carbon modeling including dynamic aggregation.
\newblock \emph{Journal of Plant Nutrition and Soil Science}, 187:\penalty0 130--142, 2 2024.
\newblock ISSN 15222624.
\newblock \doi{10.1002/jpln.202300080}.

\bibitem[Rabot et~al.(2024)Rabot, Saby, Martin, Barr{\'e}, Chenu, Cousin, Arrouays, Angers, and Bispo]{rabot2024relevance}
Eva Rabot, Nicolas~PA Saby, Manuel~P Martin, Pierre Barr{\'e}, Claire Chenu, Isabelle Cousin, Dominique Arrouays, Denis Angers, and Antonio Bispo.
\newblock Relevance of the organic carbon to clay ratio as a national soil health indicator.
\newblock \emph{Geoderma}, 443:\penalty0 116829, 2024.

\bibitem[KABONEKA et~al.(2024)KABONEKA, IRAKOZE, NDIHOKUBWAYO, NIYUNGEKO, and NTAKIYIRUTA]{kaboneka2024predicting}
Salvator KABONEKA, Willy IRAKOZE, Soter NDIHOKUBWAYO, Christophe NIYUNGEKO, and Pierre NTAKIYIRUTA.
\newblock Predicting soil cation exchange capacity (cec) from ph, percentage of clay and organic carbon: A case study on selected burundi surface soils.
\newblock 2024.

\bibitem[Lusch et~al.(2018)Lusch, Kutz, and Brunton]{Lusch2018}
Bethany Lusch, J.~Nathan Kutz, and Steven~L. Brunton.
\newblock Deep learning for universal linear embeddings of nonlinear dynamics.
\newblock \emph{Nature Communications}, 9:\penalty0 4950, 2018.
\newblock \doi{10.1038/s41467-018-07421-0}.

\bibitem[Timm et~al.(2006)Timm, Gomes, Barbosa, Reichardt, Souza, and Dynia]{timm2006neural}
Lu{\'\i}s~Carlos Timm, Daniel~Takata Gomes, Emanuel~Pimentel Barbosa, Klaus Reichardt, Manoel Dornelas~de Souza, and Jos{\'e}~Fl{\'a}vio Dynia.
\newblock Neural network and state-space models for studying relationships among soil properties.
\newblock \emph{Scientia Agricola}, 63:\penalty0 386--395, 2006.

\bibitem[Wang et~al.(2023)Wang, Chen, Hong, Hu, Peng, and Shi]{wang2023comparison}
Yu~Wang, Songchao Chen, Yongsheng Hong, Bifeng Hu, Jie Peng, and Zhou Shi.
\newblock A comparison of multiple deep learning methods for predicting soil organic carbon in southern xinjiang, china.
\newblock \emph{Computers and Electronics in Agriculture}, 212:\penalty0 108067, 2023.

\bibitem[Metrikaityt\u0117~Gudel\u0117 et~al.(2025)Metrikaityt\u0117~Gudel\u0117, Su\u017eiedelyt\u0117-Visockien\u0117, \v{C}erniauskait\u0117, Lampickas, and \u{S}ikarska]{Metrikaityte2025}
Giedr\u0117 Metrikaityt\u0117~Gudel\u0117, J\u016bras Su\u017eiedelyt\u0117-Visockien\u0117, Karolina \v{C}erniauskait\u0117, Aurelijus Lampickas, and Natalija \u{S}ikarska.
\newblock Quantifying soil carbon sequestration potential through carbon farming practices with {RothC} model adapted to {Lithuania}.
\newblock \emph{Land}, 14\penalty0 (7):\penalty0 1497, 2025.
\newblock \doi{10.3390/land14071497}.

\bibitem[Paul et~al.(2020)Paul, Ranaivoson, Giller, et~al.]{paul2020evaluation}
B.~K. Paul, L.~Ranaivoson, K.~E. Giller, et~al.
\newblock Evaluation of rothc and century models to simulate soil organic carbon in long-term agroecosystem experiments in the tropics.
\newblock \emph{Geoderma}, 368:\penalty0 114288, 2020.

\bibitem[Schimel(2023)]{schimel2023modeling}
Joshua Schimel.
\newblock Modeling ecosystem-scale carbon dynamics in soil: The microbial dimension.
\newblock \emph{Soil Biology and Biochemistry}, 178:\penalty0 108948, 2023.

\bibitem[Rackauckas et~al.(2020)Rackauckas, Ma, Martensen, Warner, Zubov, Supekar, Skinner, Ramadhan, and Edelman]{rackauckas2020universal}
Christopher Rackauckas, Yingbo Ma, Julius Martensen, Collin Warner, Kirill Zubov, Rohit Supekar, Dominic Skinner, Ali Ramadhan, and Alan Edelman.
\newblock Universal differential equations for scientific machine learning.
\newblock \emph{arXiv preprint arXiv:2001.04385}, 2020.

\bibitem[Raissi et~al.(2019{\natexlab{a}})Raissi, Perdikaris, and Karniadakis]{raissi2019pinn}
Maziar Raissi, Paris Perdikaris, and George~E Karniadakis.
\newblock Physics-informed neural networks: A deep learning framework for solving forward and inverse problems involving nonlinear pdes.
\newblock \emph{Journal of Computational Physics}, 378:\penalty0 686--707, 2019{\natexlab{a}}.

\bibitem[Bolibar et~al.(2023{\natexlab{a}})Bolibar, Sapienza, Maussion, Lguensat, Wouters, and P{\'e}rez]{bolibar2023glacier}
Jordi Bolibar, Facundo Sapienza, Fabien Maussion, Redouane Lguensat, Bert Wouters, and Fernando P{\'e}rez.
\newblock Universal differential equations for glacier ice flow modelling.
\newblock \emph{Geoscientific Model Development Discussions}, pages 1--26, 2023{\natexlab{a}}.

\bibitem[Chen et~al.(2018{\natexlab{a}})Chen, Rubanova, Bettencourt, and Duvenaud]{chen2018neuralODE}
Ricky~TQ Chen, Yulia Rubanova, Jesse Bettencourt, and David~K Duvenaud.
\newblock Neural ordinary differential equations.
\newblock \emph{Advances in Neural Information Processing Systems}, 31, 2018{\natexlab{a}}.

\bibitem[Dandekar(2022)]{dandekar2022phd}
Raj~Abhijit Dandekar.
\newblock \emph{A new way to do epidemic modeling}.
\newblock PhD thesis, Massachusetts Institute of Technology, 2022.

\bibitem[Baker et~al.(2019)Baker, Alexander, Bremer, Hagberg, Kevrekidis, Najm, Parashar, Patra, Sethian, Wild, et~al.]{baker2019workshop}
Nathan Baker, Frank Alexander, Timo Bremer, Aric Hagberg, Yannis Kevrekidis, Habib Najm, Manish Parashar, Abani Patra, James Sethian, Stefan Wild, et~al.
\newblock Workshop report on basic research needs for scientific machine learning: Core technologies for artificial intelligence.
\newblock Technical report, USDOE Office of Science (SC), Washington, DC (United States), 2019.

\bibitem[Verma et~al.(2024)Verma, Heinonen, and Garg]{Verma2024}
Yogesh Verma, Markus Heinonen, and Vikas Garg.
\newblock Climode: Climate and weather forecasting with physics-informed neural odes.
\newblock \emph{arXiv preprint arXiv:2404.10024}, 2024.

\bibitem[Cuomo et~al.(2022)Cuomo, Di~Cola, Cutr\`{e}, and Piccialli]{Cuomo2022}
Salvatore Cuomo, Valeria~Sa Di~Cola, Giulia Cutr\`{e}, and Francesco Piccialli.
\newblock Scientific machine learning through physics-informed neural networks: Where we are and what's next.
\newblock \emph{Journal of Scientific Computing}, 92:\penalty0 88, 2022.

\bibitem[Cai et~al.(2021)Cai, Mao, Wang, Yin, and Karniadakis]{cai2021pinnfluid}
Shengze Cai, Zhiping Mao, Zhicheng Wang, Minglang Yin, and George~Em Karniadakis.
\newblock Physics-informed neural networks (pinns) for fluid mechanics: A review.
\newblock \emph{Acta Mechanica Sinica}, 37\penalty0 (12):\penalty0 1727--1738, 2021.

\bibitem[Karniadakis et~al.(2021)Karniadakis, Kevrekidis, Lu, Perdikaris, Wang, and Yang]{karniadakis2021pinnreview}
George~Em Karniadakis, Ioannis~G Kevrekidis, Lu~Lu, Paris Perdikaris, Sifan Wang, and Liu Yang.
\newblock Physics-informed machine learning.
\newblock \emph{Nature Reviews Physics}, 3\penalty0 (6):\penalty0 422--440, 2021.

\bibitem[Hossain et~al.(2024)Hossain, Fanfani, Fischer, and Quackenbush]{Hossain2024}
Intekhab Hossain, Viola Fanfani, Jonas Fischer, and John Quackenbush.
\newblock Biologically informed neuralodes for genome-wide regulatory dynamics.
\newblock \emph{Genome Biology}, 25:\penalty0 127, 2024.
\newblock \doi{10.1186/s13059-024-03264-0}.

\bibitem[Bonnaff\'{e} et~al.(2021)Bonnaff\'{e}, Sheldon, and Coulson]{Bonnaffe2021}
William Bonnaff\'{e}, Ben~C. Sheldon, and Tim Coulson.
\newblock Neural ordinary differential equations for ecological and evolutionary time-series analysis.
\newblock \emph{Methods in Ecology and Evolution}, 12\penalty0 (7):\penalty0 1301--1315, 2021.
\newblock \doi{10.1111/2041-210X.13606}.

\bibitem[Vogel et~al.(2024)Vogel, Amelung, Baum, Bonkowski, Blagodatsky, Grosch, Herbst, Kiese, Koch, Kuhwald, et~al.]{vogel2024adequately}
H-J Vogel, Wulf Amelung, C~Baum, Michael Bonkowski, Sergey Blagodatsky, R~Grosch, M~Herbst, R~Kiese, S~Koch, M~Kuhwald, et~al.
\newblock How to adequately represent biological processes in modeling multifunctionality of arable soils.
\newblock \emph{Biology and Fertility of Soils}, 60\penalty0 (3):\penalty0 263--306, 2024.

\bibitem[Chen et~al.(2018{\natexlab{b}})Chen, Rubanova, Bettencourt, and Duvenaud]{Chen2018}
Tian~Qi Chen, Yulia Rubanova, Jesse Bettencourt, and David Duvenaud.
\newblock Neural ordinary differential equations.
\newblock In \emph{Advances in Neural Information Processing Systems (NeurIPS)}, pages 6571--6583, 2018{\natexlab{b}}.

\bibitem[Kidger(2022)]{Kidger2022}
Patrick Kidger.
\newblock On neural differential equations.
\newblock \emph{Journal of Computational Dynamics}, 9\penalty0 (2):\penalty0 191--212, 2022.

\bibitem[Zoppou and Knight(1997)]{zoppou1997analytical}
Christopher Zoppou and JH~Knight.
\newblock Analytical solutions for advection and advection-diffusion equations with spatially variable coefficients.
\newblock \emph{Journal of Hydraulic Engineering}, 123\penalty0 (2):\penalty0 144--148, 1997.

\bibitem[Jiwari(2015)]{jiwari2015hybrid}
Ram Jiwari.
\newblock A hybrid numerical scheme for the numerical solution of the burgers’ equation.
\newblock \emph{Computer Physics Communications}, 188:\penalty0 59--67, 2015.

\bibitem[Jadon et~al.(2024)Jadon, Patil, and Jadon]{jadon2024regressionloss}
A.~Jadon, A.~Patil, and S.~Jadon.
\newblock A comprehensive survey of regression-based loss functions for time series forecasting.
\newblock In \emph{Proceedings of the International Conference on Data Management, Analytics \& Innovation}, pages 117--147, Singapore, January 2024. Springer Nature Singapore.
\newblock \doi{10.1007/978-981-99-7870-5\_10}.

\bibitem[Sancho et~al.(1982)Sancho, San~Miguel, Katz, and Gunton]{sancho1982multiplicative}
J.~M. Sancho, M.~San~Miguel, S.~L. Katz, and J.~D. Gunton.
\newblock Analytical and numerical studies of multiplicative noise.
\newblock \emph{Physical Review A}, 26\penalty0 (3):\penalty0 1589--1609, 1982.
\newblock \doi{10.1103/PhysRevA.26.1589}.

\bibitem[Ermolieva et~al.(2024{\natexlab{b}})Ermolieva, Havlik, Lessa-Derci-Augustynczik, Frank, Balkovic, Skalsky, Deppermann, Nakhavali, Komendantova, Kahil, Wang, Folberth, and Knopov]{Ermolieva2024}
Tatiana Ermolieva, Petr Havlik, Andrey Lessa-Derci-Augustynczik, Stefan Frank, Juraj Balkovic, Rastislav Skalsky, Andre Deppermann, Mahdi Nakhavali, Nadejda Komendantova, Taher Kahil, Gang Wang, Christian Folberth, and Pavel~S. Knopov.
\newblock Tracking the dynamics and uncertainties of soil organic carbon in agricultural soils based on a novel robust meta-model framework using multisource data.
\newblock \emph{Sustainability (Switzerland)}, 16, 8 2024{\natexlab{b}}.
\newblock ISSN 20711050.
\newblock \doi{10.3390/su16166849}.

\bibitem[Fu et~al.(2024)Fu, Clanton, Demuth, Goodman, Griffith, Khim-Young, Maddalena, LaMarca, Wright, Schurman, and Kellner]{Fu2024}
Peng Fu, Christian Clanton, Kirk~M. Demuth, Verena Goodman, Lauren Griffith, Mage Khim-Young, Julia Maddalena, Kenny LaMarca, Logan~A. Wright, David~W. Schurman, and James~R. Kellner.
\newblock Accurate quantification of 0–30 cm soil organic carbon in croplands over the continental united states using machine learning.
\newblock \emph{Remote Sensing}, 16, 6 2024.
\newblock ISSN 20724292.
\newblock \doi{10.3390/rs16122217}.

\bibitem[Zhang et~al.(2024)Zhang, Heuvelink, Mulder, Chen, Deng, and Yang]{Zhang2024}
Lei Zhang, Gerard~B.M. Heuvelink, Vera~L. Mulder, Songchao Chen, Xunfei Deng, and Lin Yang.
\newblock Using process-oriented model output to enhance machine learning-based soil organic carbon prediction in space and time.
\newblock \emph{Science of the Total Environment}, 922, 4 2024.
\newblock ISSN 18791026.
\newblock \doi{10.1016/j.scitotenv.2024.170778}.

\bibitem[Emadi et~al.(2020)Emadi, Taghizadeh-Mehrjardi, Cherati, Danesh, Mosavi, and Scholten]{Emadi2020}
Mostafa Emadi, Ruhollah Taghizadeh-Mehrjardi, Ali Cherati, Majid Danesh, Amir Mosavi, and Thomas Scholten.
\newblock Predicting and mapping of soil organic carbon using machine learning algorithms in northern iran.
\newblock 2020.

\bibitem[Cotrufo and Lavallee(2022)]{Cotrufo2022}
M.~Francesca Cotrufo and Jocelyn~M. Lavallee.
\newblock \emph{Soil organic matter formation, persistence, and functioning: A synthesis of current understanding to inform its conservation and regeneration}, volume 172, pages 1--66.
\newblock Academic Press Inc., 1 2022.
\newblock ISBN 9780323989534.
\newblock \doi{10.1016/bs.agron.2021.11.002}.

\bibitem[Pierson et~al.(2022)Pierson, Lohse, Wieder, Patton, Facer, de~Graaff, Georgiou, Seyfried, Flerchinger, and Will]{Pierson2022}
Derek Pierson, Kathleen~A. Lohse, William~R. Wieder, Nicholas~R. Patton, Jeremy Facer, Marie~Anne de~Graaff, Katerina Georgiou, Mark~S. Seyfried, Gerald Flerchinger, and Ryan Will.
\newblock Optimizing process-based models to predict current and future soil organic carbon stocks at high-resolution.
\newblock \emph{Scientific Reports}, 12, 12 2022.
\newblock ISSN 20452322.
\newblock \doi{10.1038/s41598-022-14224-8}.

\bibitem[Zhou et~al.(2024)Zhou, Ren, Wang, Delgado-Baquerizo, Luo, Luo, Du, Zhu, Yang, Jiao, Zhao, Cai, Yang, and Wei]{Zhou2024}
Zhenghu Zhou, Chengjie Ren, Chuankuan Wang, Manuel Delgado-Baquerizo, Yiqi Luo, Zhongkui Luo, Zhenggang Du, Biao Zhu, Yuanhe Yang, Shuo Jiao, Fazhu Zhao, Andong Cai, Gaihe Yang, and Gehong Wei.
\newblock Global turnover of soil mineral-associated and particulate organic carbon.
\newblock \emph{Nature Communications}, 15, 12 2024.
\newblock ISSN 20411723.
\newblock \doi{10.1038/s41467-024-49743-7}.

\bibitem[Broek et~al.(2025)Broek, Govers, Schrumpf, and Six]{VanDeBroek2025}
Marijn Van~De Broek, Gerard Govers, Marion Schrumpf, and Johan Six.
\newblock A microbially driven and depth-explicit soil organic carbon model constrained by carbon isotopes to reduce parameter equifinality.
\newblock \emph{Biogeosciences}, 22:\penalty0 1427--1446, 3 2025.
\newblock ISSN 17264189.
\newblock \doi{10.5194/bg-22-1427-2025}.

\bibitem[Chen et~al.(2023)Chen, Du, Weng, Sun, Zhang, Liu, Yang, Li, Wang, Luo, Gao, Chen, Pan, and Van Zwieten]{Chen2023}
Yalan Chen, Zhangliu Du, Zhe Weng, Ke~Sun, Yuqin Zhang, Qin Liu, Yan Yang, Yang Li, Zhibo Wang, Yu~Luo, Bo~Gao, Bin Chen, Zezhen Pan, and Lukas Van Zwieten.
\newblock Formation of soil organic carbon pool is regulated by the structure of dissolved organic matter and microbial carbon pump efficacy: A decadal study comparing different carbon management strategies.
\newblock \emph{Global Change Biology}, 29:\penalty0 5445--5459, 9 2023.
\newblock ISSN 13652486.
\newblock \doi{10.1111/gcb.16865}.

\bibitem[Zhao et~al.(2023)Zhao, Mao, Gao, Lu, Pan, and Li]{Zhao2023}
Zhanhui Zhao, Yanli Mao, Songfeng Gao, Chunyang Lu, Chuanjiao Pan, and Xiaoyu Li.
\newblock Organic carbon accumulation and aggregate formation in soils under organic and inorganic fertilizer management practices in a rice–wheat cropping system.
\newblock \emph{Scientific Reports}, 13, 12 2023.
\newblock ISSN 20452322.
\newblock \doi{10.1038/s41598-023-30541-y}.

\bibitem[Pavlovic et~al.(2024{\natexlab{b}})Pavlovic, Ilic, Ralevic, Antonic, Raffa, Bandecchi, and Culibrk]{Pavlovic2024}
Marko Pavlovic, Slobodan Ilic, Neobojša Ralevic, Nenad Antonic, Dylan~Warren Raffa, Michele Bandecchi, and Dubravko Culibrk.
\newblock A deep learning approach to estimate soil organic carbon from remote sensing.
\newblock \emph{Remote Sensing}, 16, 2 2024{\natexlab{b}}.
\newblock ISSN 20724292.
\newblock \doi{10.3390/rs16040655}.

\bibitem[Si et~al.(2023)Si, Chen, Wei, Zhang, Sun, and Liang]{Si2023}
Qintana Si, Kangli Chen, Bin Wei, Yaowen Zhang, Xun Sun, and Junyi Liang.
\newblock Formation of particulate organic carbon from dissolved substrate input enhances soil carbon sequestration, 7 2023.
\newblock URL \url{https://egusphere.copernicus.org/preprints/2023/egusphere-2023-1483/}.

\bibitem[Wieder et~al.(2014)Wieder, Grandy, Kallenbach, and Bonan]{Wieder2014}
W.~R. Wieder, A.~S. Grandy, C.~M. Kallenbach, and G.~B. Bonan.
\newblock Integrating microbial physiology and physio-chemical principles in soils with the microbial-mineral carbon stabilization (mimics) model.
\newblock \emph{Biogeosciences}, 11:\penalty0 3899--3917, 7 2014.
\newblock ISSN 17264189.
\newblock \doi{10.5194/bg-11-3899-2014}.

\bibitem[Bai et~al.(2014)Bai, Blumfield, and Reverchon]{HosseiniBai2014}
Shahla~Hosseini Bai, Timothy~J. Blumfield, and Frédérique Reverchon.
\newblock The impact of mulch type on soil organic carbon and nitrogen pools in a sloping site.
\newblock \emph{Biology and Fertility of Soils}, 50:\penalty0 37--44, 1 2014.
\newblock ISSN 01782762.
\newblock \doi{10.1007/s00374-013-0829-z}.

\bibitem[Jia et~al.(2021)Jia, Jacques, Gérard, Su, Mayer, and Šimůnek]{Jia2021}
M.~Jia, D.~Jacques, F.~Gérard, D.~Su, K.~U. Mayer, and J.~Šimůnek.
\newblock A benchmark for soil organic matter degradation under variably saturated flow conditions.
\newblock \emph{Computational Geosciences}, 25:\penalty0 1359--1377, 8 2021.
\newblock ISSN 15731499.
\newblock \doi{10.1007/s10596-019-09862-3}.

\bibitem[Zhang et~al.(2020)Zhang, Goll, Wang, Ciais, Wieder, Abramoff, Huang, Guenet, Prescher, Rossel, Barré, Chenu, Zhou, and Tang]{Zhang2020}
Haicheng Zhang, Daniel~S. Goll, Ying~Ping Wang, Philippe Ciais, William~R. Wieder, Rose Abramoff, Yuanyuan Huang, Bertrand Guenet, Anne~Katrin Prescher, Raphael A.~Viscarra Rossel, Pierre Barré, Claire Chenu, Guoyi Zhou, and Xuli Tang.
\newblock Microbial dynamics and soil physicochemical properties explain large-scale variations in soil organic carbon.
\newblock \emph{Global Change Biology}, 26:\penalty0 2668--2685, 4 2020.
\newblock ISSN 13652486.
\newblock \doi{10.1111/gcb.14994}.

\bibitem[Yu et~al.(2020)Yu, Ahrens, Wutzler, Schrumpf, and Zaehle]{Yu2020}
Lin Yu, Bernhard Ahrens, Thomas Wutzler, Marion Schrumpf, and Sonke Zaehle.
\newblock Jena soil model (jsm v1.0; revision 1934): A microbial soil organic carbon model integrated with nitrogen and phosphorus processes.
\newblock \emph{Geoscientific Model Development}, 13:\penalty0 783--803, 2 2020.
\newblock ISSN 19919603.
\newblock \doi{10.5194/gmd-13-783-2020}.

\bibitem[Fabrizzi et~al.(2009)Fabrizzi, Rice, Amado, Fiorin, Barbagelata, and Melchiori]{Fabrizzi2009}
Karina~P. Fabrizzi, Charles~W. Rice, Telmo~J.C. Amado, Jackson Fiorin, Pedro Barbagelata, and Ricardo Melchiori.
\newblock Protection of soil organic c and n in temperate and tropical soils: Effect of native and agroecosystems.
\newblock In \emph{Biogeochemistry}, volume~92, pages 129--143, 1 2009.
\newblock \doi{10.1007/s10533-008-9261-0}.

\bibitem[()]{}
\emph{Global change effects on microbial-mediated soil organic carbon cycling processes}.
\newblock PhD thesis, Vrije Universiteit Amsterdam, 6 2024.
\newblock URL \url{https://hdl.handle.net/1871.1/521f8989-5e01-4d26-89d4-6680751bd2ec}.

\bibitem[Fan et~al.(2021)Fan, Gao, Zhao, Wang, Qu, Zhang, and Bai]{Fan2021}
Xianlei Fan, Decai Gao, Chunhong Zhao, Chao Wang, Ying Qu, Jing Zhang, and Edith Bai.
\newblock Improved model simulation of soil carbon cycling by representing the microbially derived organic carbon pool.
\newblock \emph{ISME Journal}, 15:\penalty0 2248--2263, 8 2021.
\newblock ISSN 17517370.
\newblock \doi{10.1038/s41396-021-00914-0}.

\bibitem[Shen et~al.(2022)Shen, Ramirez-Lopez, Behrens, Cui, Zhang, Walden, Wetterlind, Shi, Sudduth, Baumann, Song, Catambay, and Rossel]{Shen2022}
Zefang Shen, Leonardo Ramirez-Lopez, Thorsten Behrens, Lei Cui, Mingxi Zhang, Lewis Walden, Johanna Wetterlind, Zhou Shi, Kenneth~A. Sudduth, Philipp Baumann, Yongze Song, Kevin Catambay, and Raphael A.~Viscarra Rossel.
\newblock Deep transfer learning of global spectra for local soil carbon monitoring.
\newblock \emph{ISPRS Journal of Photogrammetry and Remote Sensing}, 188:\penalty0 190--200, 6 2022.
\newblock ISSN 09242716.
\newblock \doi{10.1016/j.isprsjprs.2022.04.009}.

\bibitem[Bursać et~al.(2022)Bursać, Kovačević, and Bajat]{Bursa2022}
Petar Bursać, Miloš Kovačević, and Branislav Bajat.
\newblock Instance-based transfer learning for soil organic carbon estimation.
\newblock \emph{Frontiers in Environmental Science}, 10, 9 2022.
\newblock ISSN 2296665X.
\newblock \doi{10.3389/fenvs.2022.1003918}.

\bibitem[Liu et~al.(2012)Liu, Shao, and Wang]{liu2012estimating}
Zhi-Peng Liu, Ming-An Shao, and Yun-Qiang Wang.
\newblock Estimating soil organic carbon across a large-scale region: a state-space modeling approach.
\newblock \emph{Soil science}, 177\penalty0 (10):\penalty0 607--618, 2012.

\bibitem[Raissi et~al.(2019{\natexlab{b}})Raissi, Perdikaris, and Karniadakis]{Raissi2019}
Maziar Raissi, Paris Perdikaris, and George~E. Karniadakis.
\newblock Physics-informed neural networks: A deep learning framework for solving forward and inverse problems involving nonlinear partial differential equations.
\newblock \emph{Journal of Computational Physics}, 378:\penalty0 686--707, 2019{\natexlab{b}}.

\bibitem[Bolibar et~al.(2023{\natexlab{b}})Bolibar, Sapienza, Maussion, Lguensat, Wouters, and P\'{e}rez]{Bolibar2023}
Jordi Bolibar, Facundo Sapienza, Fabien Maussion, Redouane Lguensat, Bert Wouters, and Fernando P\'{e}rez.
\newblock Universal differential equations for glacier ice flow modelling.
\newblock \emph{Geoscientific Model Development}, 16:\penalty0 6671--6687, 2023{\natexlab{b}}.
\newblock \doi{10.5194/gmd-16-6671-2023}.

\end{thebibliography}
\nocite*{}

\end{document}